# MedNet-PVS: A MedNeXt-Based Deep Learning Model for Automated Segmentation of Perivascular Spaces

Zhen Xuen Brandon Low[1*], Rory Zhang[2], Hang Min[3,4], William Pham[5], Lucy Vivash[5,6], Jasmine Moses[5], Miranda Lynch[5], Karina Dorfman[5], Cassandra Marotta[5], Shaun Koh[5], Jacob Bunyamin[5], Ella Rowsthorn[5], Alex Jarema[7], Himashi Peiris[8,9], Zhaolin Chen[8,9], Sandy R Shultz[5,6], David K Wright[5], Dexiao Kong[10,11,12], Sharon L. Naismith[10,11,12], Terence J. O'Brien[5,6], Ying Xia[3], Meng Law[5,7], Benjamin Sinclair[5]

[1] Jeffrey Cheah School of Medicine and Health Sciences, Monash University Malaysia, Selangor, Malaysia

[2] Melbourne Bioinnovation Student Initiative (MBSI), Parkville, VIC, Australia

[3] CSIRO Australian e-Health Research Centre, Herston, Queensland, Australia

[4] South Western Clinical School, University of New South Wales, Sydney, Australia

[5] Department of Neuroscience, The School of Translational Medicine, Monash University, Melbourne, Australia

[6] Department of Neurology, Alfred Health, Melbourne, Australia

[7] Department of Radiology, Alfred Health, Melbourne, Australia

[8] Monash Biomedical Imaging. Monash University, Melbourne, Australia

[9] Department of Data Science and AI, Monash University, Melbourne, Australia

[10] School of Psychology, Faculty of Science, University of Sydney, Sydney, Australia

[11] Healthy Brain Ageing Program, Brain and Mind Centre, University of Sydney, Camperdown, NSW, Australia

[12] Charles Perkins Centre, University of Sydney, Camperdown, NSW, Australia

* **Corresponding author:** Zhen Xuen Brandon Low**,** Jeffrey Cheah School of Medicine and Health Sciences, Monash University Malaysia, Selangor, Malaysia. **E-mail address:** zlow0014@student.monash.edu

## ABSTRACT

Perivascular spaces (PVS) are fluid-filled structures surrounding cerebral blood vessels that play a key role in the brain's waste clearance system. Enlarged PVS are increasingly recognized as biomarkers of cerebral small vessel disease, Alzheimer's disease, stroke, and aging-related neurodegeneration. However, manual segmentation of PVS is time-consuming and subject to moderate inter-rater reliability, while existing automated deep learning models have moderate performance, and typically fail to generalize across diverse clinical and research MRI datasets. We adapted MedNeXt-L-k5, a Transformer-inspired 3D encoder-decoder convolutional network, for automated PVS segmentation. Two models were trained: one using a homogeneous dataset of 200 T2-weighted (T2w) MRI scans from the Human Connectome Project-Aging (HCP-Aging) dataset and another using 40 heterogeneous T1-weighted (T1w) MRI volumes from seven studies across six scanners, including healthy individuals, mild cognitive impairment, and dementia patients. Model performance was evaluated using internal 5-fold cross validation (5FCV) and leave-one-site-out cross validation (LOSOCV). MedNeXt-L-k5 models trained on the T2w images of the HCP-Aging dataset achieved competitive voxel-level dice scores of 0.88±0.06 (white matter, WM), comparable to the reported inter-rater reliability of that dataset, and the highest yet reported in the literature. The same models trained on the T1w images of the HCP-Aging dataset achieved a substantially lower dice score of 0.58±0.09 (WM). This was a similar performance to that of the MedNeXt-L-k5 models trained on the heterogeneous 7-site T1w dataset, achieving voxel-level Dice scores of 0.51±0.14 (WM) and 0.53±0.11 (basal ganglia, BG), and cluster-level Dice scores of 0.64±0.17 (WM) and 0.65±0.16 (BG). Under stringent LOSOCV, the model had voxel-level Dice scores of 0.38±0.16 (WM) and 0.35±0.12 (BG), and cluster-level Dice scores of 0.61±0.19 (WM) and 0.62±0.21 (BG). MedNeXt-L-k5 provides an efficient, reliable solution for automated PVS segmentation across diverse T1w and T2w MRI datasets. Performances on T2w images were far higher than on T1w images. MedNeXt-L-k5 did not outperform the nnU-Net, indicating that the attention-based mechanisms present in transformer-inspired models to provide global context are not required for high accuracy in PVS segmentation.

## 1. INTRODUCTION

Perivascular spaces (PVS), also known as Virchow-Robin spaces, are fluid-filled compartments that surround cerebral blood vessels and play a critical role in the brain's waste clearance system. These spaces have gained increasing attention over the past decade, largely driven by advancements in *in vivo* neuroimaging techniques. Enlarged PVS have been identified normal aging (Laveskog et al., 2018) and a range of neurological and systemic conditions, including Alzheimer's disease (Lynch et al., 2022), cerebral small vessel disease (Gyanwali et al., 2019), autism spectrum disorder (Sotgiu et al., 2023), depression (Ranti et al., 2022), sleep disorders (Aribisala et al., 2023). In acute stroke, PVS enlargement has been associated with greater cerebral edema severity (Mestre et al., 2020), and in patients with prior ischemic stroke, it has been linked to an increased risk of haemorrhagic stroke (Tian et al., 2024). Moreover, impaired PVS function may hinder the clearance of metabolic waste products, such as amyloid-$\beta$ (A$\beta$) plaques (Perosa et al., 2022) and tau protein aggregates (Vilor-Tejedor et al., 2021), thereby contributing to neuronal damage and cognitive decline.

Given their clinical relevance, PVS have emerged as valuable neuroimaging biomarkers for early diagnosis and disease monitoring. Traditionally, enlarged PVS are identified and manually counted on representative magnetic resonance imaging (MRI) slices to quantify PVS burden (Potter et al., 2015). However, this manual approach is labour-intensive, prone to inter-observer variability, and limited in scalability. More recently, as high-field MRI has improved PVS visualization, the need for reliable, automated segmentation methods has grown. To address this, several automated segmentation algorithms have been developed, including filtering-based techniques (Ranti et al., 2022; Duarte Coello et al., 2024), machine learning, and deep learning approaches (Lan et al., 2023; Rashid et al., 2023; Sinclair et al., 2024; Spijkerman et al., 2022).

Recent advances have seen Transformer-based architectures become increasingly popular in medical image segmentation because of their ability to model long-range dependencies and capture global context (Kumar, 2025). This capability is likely important for PVS segmentation, where accurate delineation requires the integration of both fine local details, such as shape and intensity, and broader contextual patterns (Boutinaud et al., 2021), such as anatomical location and contrast to noise of PVS compared to other similar structures. Despite these advantages, the widespread application of Transformers is limited by their dependency on large, annotated datasets with high-quality labels, which are often scarce in clinical settings (He et al., 2023; Wang et al., 2025).

Considering these limitations, we investigate MedNeXt (Roy et al., 2023), a 3D ConvNeXt-based architecture customized to the challenges of data-scarce medical settings and built upon ConvNeXt (Liu et al., 2022), which was recently introduced to re-establish the competitive performance of convolutional networks for natural images by retaining the inherent inductive bias of convolutions while incorporating architectural improvements inspired by Transformers. Its use of large, scalable kernels and residual upsampling/downsampling blocks enables effective capture of both local details and global context. We hypothesize that these features may improve segmentation performance for PVS, as accurate identification of these small, elongated, and sometimes low-contrast structures likely benefits from integrating information across multiple spatial scales.

In this work, we evaluated MedNeXt (Roy et al., 2023) for PVS segmentation across both T1-weighted (T1w) and T2-weighted (T2w) MRI datasets, spanning a diverse range of imaging protocols, scanners, and participant demographics. We trained on a large T2w and T1w dataset ($n = 200$) of cognitively healthy adults from the Human Connectome Project-Aging (HCP-A) project. To rigorously assess generalizability, we then trained and validated on a heterogeneous, multi-centre T1w dataset ($n = 40$). MedNeXt was designed to accommodate variability across imaging sites and subject populations, and prior work has demonstrated its competitive performance in medical image segmentation tasks. Here, we extend its application to PVS segmentation in both T1w and T2w modalities, with the goal of developing a robust and generalizable tool for clinical and research use.

## 2. METHODS

### 2.1.1 HCP-Aging MRI Dataset with Manual Segmentations

To evaluate the performance of MedNeXt on T2-weighted MRI, we utilized 3D T2-weighted images from the publicly available Human Connectome Project-Aging (HCP-Aging) Lifespan Release 2.0 (Bookheimer et al., 2019), with white matter PVS segmentations provided by Chai et al. (2025). This dataset comprised 200 cognitively healthy subjects aged 36 to over 100 years. At recruitment, participants were in good general health, free of diagnosed neurologic or major psychiatric disorders, symptomatic stroke, or Alzheimer's disease, with Montreal Cognitive Assessment (MoCA) scores $\geq 20$. Chai et al. started from an initial pool of 765 subjects, of which 217 were selected after balancing age and sex distribution and excluding 17 scans with motion artifacts or large lacunar infarcts via visual inspection of T2-weighted images (MRI acquisition parameters listed in Table S1). Basal ganglia PVS segmentations were not included due to their non-availability in this dataset.

White matter (WM) PVS segmentations were obtained directly from the dataset provided by Chai et al. (2025), who combined manual and automated methods to generate high-quality labels. Image preprocessing steps, including applying brain masks (derived and registered from T1-weighted images) to exclude non-brain tissues, followed by histogram equalization and non-local mean denoising, were applied only during the process of generating ground truth segmentations, as detailed in Chai et al. (2025). These steps were not applied to the T2w images used as model inputs during training and evaluation. To generate the ground truth masks, Chai et al. performed initial auto-segmentation using a Frangi filter with small sigma values to detect tubular PVS (1-2 voxels in diameter), restricted to white matter via an MNI-CIVET mask. For 40 subjects spanning all age decades, two neuroanatomists manually refined these outputs in ITK-SNAP (Yushkevich et al., 2006), correcting errors and excluding deep gray matter PVS, guided by tubular morphology across three orthogonal slices. These refinements were used to train a U-Net model (Çiçek et al., 2016) in a semi-supervised framework, which segmented PVS in the remaining 160 subjects. A team of two neuroradiologists and one trained medical student then manually reviewed and edited these segmentations, resolving uncertainties through consensus.

*2.1.2 Intra-Subject T1w-T2w Registration of HCP-Aging Images*

To a) develop a T1w PVS segmentation algorithm and b) directly compare the MedNeXt-L-k5 models trained on the HCP-Aging dataset (*n* = 200) to our own in-house dataset (*n* = 40), we registered the 200 T1w images from the HCP-Aging dataset to their corresponding T2w images, for which WM PVS ground truth segmentations are available.

Registration was performed using the Linear Image Registration Tool (FLIRT) from the FMRIB Software Library (FSL). A rigid-body transformation (6 degrees of freedom: 3 translations and 3 rotations) was used to account for intra-subject head movement. Mutual information was used as the cost function to optimize alignment across modalities (T1w and T2w). Following registration, each T1w image was resampled into the native T2w space using B-spline interpolation to preserve structural detail and minimize interpolation artifacts.

## 2.2 In-house T1w PVS Segmentation

PVS on T1w imaging were manually labelled using ITK-SNAP (version 3.3.0) (Yushkevich et al., 2006). Seven in-house raters (JM, ML1, WP, KD, CM, SK, JB) received training from an experienced neuroradiologist (ML2) and an imaging scientist (BS) to identify PVS and differentiate them from similar imaging features, such as lacunes, white matter hyperintensities, and sulci. For each cohort, one rater segmented all voxels deemed part of PVS, followed by a second rater who reviewed the segmentations, adding missed voxels and flagging discrepancies. Disagreements were resolved by a senior neuroradiologist (ML2). For the two largest datasets, ADNI and AF, all segmentations underwent additional review by a senior rater (ML2) to ensure consistency and accuracy across cohorts.

## 2.3 T1w Pre-processing

Pre-processing of the in-house T1w dataset was explored to enhance segmentation performance by minimizing inter-subject variations and improving contrast. Drawing from the work of Huang et al. (2024), which emphasized the impact of image contrast on the visibility and annotation of perivascular spaces (PVS), a similar set of pre-processing steps to standardize the input data and improve contrast was trialled. The steps include Brain Extraction with SPM12, bias field correction with ANTs, and Fuzzy-C-Means intensity normalisation. This preprocessing reduced segmentation performance (Results section 3.5), so we did not test the same preprocessing steps on the T2w data.

*2.4 MedNeXt Preprocessing and Data Augmentation*

MedNeXt leverages the nnU-Net version 1 (Isensee et al., 2021) pre-processing framework. In addition to the default pre-processing pipeline, the target spacing (resampled resolution) was manually set to 0.8 mm$^3$, the highest resolution across the datasets, rather than allowing the nnU-Net to automatically determine the median resolution. This adjustment was necessary because PVS are fine structures that are often at the resolution limit of MRI. Both T1- and T2-weighted images were intensity-normalized to a z-score and resampled to the specified target spacing. During training, data augmentation techniques such as rotations, scaling, Gaussian noise, Gaussian blur, brightness and contrast adjustments, low-resolution simulation, gamma correction, and mirroring were applied.

*2.5.1 Model Architecture*

MedNeXt is a state-of-the-art, Transformer-inspired, fully ConvNeXt 3D Encoder-Decoder Network specifically designed for medical image segmentation. MedNeXt modernizes the conventional ConvNeXt architecture by integrating key innovations such as large kernel depthwise convolutions, residual inverted bottlenecks in both upsampling and downsampling stages, and a novel weight initialization technique known as UpKern. These design choices enable the network to effectively capture fine local details and long-range spatial dependencies, while offering scalability in depth, width, and kernel size. MedNeXt is available in four different configurations—small (S), base (B), medium (M), and large (L)—which differ in the number of ConvNeXt-based up and downsampling blocks and the expansion ratio. For our experiments, we used the MedNeXt-L configuration with a 5x5x5 kernel (hereafter referred to as MedNeXt-L-k5), providing increased capacity and a larger receptive field that is crucial for detecting the subtle features characteristic of PVS. The benefit of enlarging the receptive field in 3D segmentation has also been demonstrated in recent work, where large kernel designs improved the capture of both local details and long-range context (Li et al., 2024).

Consistent with these reports, our preliminary experiments showed that the 5x5x5 kernel outperformed the 3x3x3 variant on our dataset.

*2.5.2 Model Training*

MedNeXt was developed by the same research group (Division of Medical Image Computing, DKFZ) behind nnU-Net, and as such, MedNeXt utilises the same training and pre-processing pipeline as nnU-Net. All models were trained from scratch. When training MedNeXt-L, we used a learning rate of 3e-4 and limited the training to 500 epochs (performance plateaued before this point), default nnU-Net settings were used for the rest of the training pipeline. We then compared the resulting model's segmentation performance with nnU-Net version 2 (Isensee et al., 2021) trained on default settings except for a manually specified target spacing of 0.8 mm$^3$. Although some papers have reported training deep-learning models on T2w images (Cai et al., 2024; Choi et al., 2020), none are publicly available for comparison.

*2.6.1 T1w Internal Evaluation*

We employed nnU-Net's default 5-fold cross-validation (5FCV) procedure for training and evaluation. This approach yields five models, each trained on a different 80% subset of the data. These models can then be combined to run inference on new data, ensuring robust and generalizable performance.

*2.6.2 T1w External Evaluation*

To determine the model's performance on images from an unseen dataset with characteristics that differ from the training data, a leave-one-site-out cross validation (LOSOCV) approach was used in the second training and evaluation phase. In this setup, the nnU-Net was trained on data from five out of six sites (ADNI, AF, HBA, FTD, MCIS) and then evaluated on the remaining site (ASC). This approach mimics real-world scenarios where the model encounters new, previously unseen data with different properties from the training set.

*2.7 Post-processing*

To evaluate PVS volumes and counts within specific regions of interest, each subject's T1-weighted (T1w) image was non-linearly registered to the MNI template space provided by FSL (Jenkinson et al., 2012) using ANTS Syn registration (Avants et al., 2008). The resulting inverse deformation fields were then employed to propagate atlas segmentations of white matter (WM) and the basal ganglia (BG). The WM segmentation was sourced from the Desikan-Kiliany cortical atlas (Desikan et al., 2006), while the BG regions, including the putamen, pallidum, caudate, nucleus accumbens, and thalamus, were segmented using FreeSurfer's automated subcortical segmentation. Additionally, the internal capsule (both anterior and posterior limbs) was extracted from the JHU DTI-based WM atlas (Mori et al., 2006), added to the BG region of interest, and the WM region of interest was defined by removing the BG region of interest (including internal capsule) from the initial WM mask. Performance metrics for the model were then calculated individually for each of these regions.

*2.8 Model Evaluation*

For each model outlined in the previous section, the primary metric used was the voxel-level Dice score ($DSC_{vox}$), which quantifies the overlap between the ground truth and the algorithm's segmentations. $DSC_{vox}$ was computed for every subject in the validation sets and then averaged across subjects. In addition to the Dice score, we evaluated voxel-level sensitivity ($Sen_{vox}$), positive predictive value ($PPV_{vox}$), and the correlation ($r_{vox}$) between the manually segmented and algorithm-derived PVS volumes. Because counts of MRI-visible PVS are often used to indicate the burden of enlarged PVS, we also calculated the count-level (or cluster-level) equivalents of these measures. Unique PVS clusters were identified by segmenting distinct connected voxels using the skimage.measure.labels algorithm from Python's skimage package. A cluster was classified as a true positive if any voxel within the algorithm-segmented cluster overlapped with the manual segmentation, and a false positive if there was no overlap. Definitions for these outcome measures are provided in Table 1.

***Table 1****: Performance measures. n denotes the number of distinct clusters available in segmentation.*

| Performance measure | Voxel Level | Cluster Level |
|---|---|---|
|  |  |  |

| Dice Score | $DSC_{vox} = \dfrac{2 * volume_{overlap}}{volume_{manual} + volume_{algo}}$ | $DSC_{num} = \dfrac{2 * n_{overlapping}}{n_{manual} + n_{algo}}$ |
|---|---|---|
| Sensitivity (Recall) | $Sen_{vox} = \dfrac{volume_{overlap}}{volume_{manual}}$ | $Sen_{num} = \dfrac{n_{overlap}}{n_{manual}}$ |
| Positive Predictive Value (Precision) | $PPV_{vox} = \dfrac{volume_{overlap}}{volume_{algo}}$ | $PPV_{num} = \dfrac{n_{overlap}}{n_{algo}}$ |
| Correlation | $r_{vox} = r(volume_{manual}, volume_{algo})$ | $r_{num} = r(n_{manual}, n_{algo})$ |

### 2.9 Comparison of HCP-Aging T2w Labels and in-house T1w Labels

To investigate whether any differences in model performance between the HCP-A T2w and in-house T1w datasets were related to underlying label attributes, we performed two complementary analyses. First, we compared the distribution of PVS cluster sizes between the two label sets to assess whether differences in performance were potentially related to an over-representation of small, difficult-to-segment PVSs, which could bias performance downward.

Second, we examined the absolute image contrast between manually labelled PVS voxels and their immediate surroundings (defined as a one-voxel dilation) in both datasets. This allowed us to test whether lower intrinsic contrast of PVSs on T1w images, relative to T2w, could contribute to any observed differences in segmentation accuracy.

### 2.10 Comparator Algorithms

To evaluate the performance of our MedNeXt-L-k5 model, we compared it against publicly available methods documented in the literature at the time of this study. A recent systematic review by Waymont et al. (2024) identified 19 deep learning algorithms developed for PVS detection as of September 2023. We conducted an updated search, identifying 6 additional algorithms not included in that review, bringing the total to 25 deep learning algorithms for PVS segmentation to date. Of these, 14 generate segmentation maps, while the remaining 11 produce alternative outputs, such as PVS rating scales or scores, distance maps, enhanced maps, or lesion heatmaps. Among the 14 segmentation algorithms, 9 are designed for 3D image processing, but only 4 were publicly available for evaluation. All four of these used T1w MRI as input.

The publicly available algorithms assessed were (i) Weakly Supervised Perivascular Space Segmentation (WPSS), which applies a Frangi filter to enhance vessel-like structures followed by a U-Net to eliminate false positives (Lan et al., 2023); (ii) SHIVA-PVS, which uses a 3D U-Net architecture (Boutinaud et al., 2021); (iii) mc-PVSnet, which leveraged the nnU-Net for training on preprocessed T1w images (Huang et al., 2024); and (iv) Perivascular space Identification Nnunet for Generalised Usage (PINGU), an nnU-Net-based model trained on the same internal T1w dataset of 40 samples used in this study (Sinclair et al., 2024). Since PINGU shares the same internal T1w dataset, we re-trained its nnU-Net using the same training/validation splits as MedNeXt-L-k5 for both the T1w and T2w datasets. The performance of WPSS, SHIVA-PVS, and mc-PVSnet was evaluated directly on our internal dataset of 40 T1w images.

## 3. RESULTS

### 3.1 Training and Evaluation on HCP-Aging Dataset

Both MedNeXt-L-k5 and nnU-Net achieved excellent segmentation performance on T2w images (Figure 1), with voxel-level Dice scores of $0.88 \pm 0.06$ and $0.89 \pm 0.09$, respectively (Figure 2, Table S3). Cluster-level Dice scores were nearly identical ($0.88 \pm 0.09$ vs. $0.89 \pm 0.09$). Both models also demonstrated high sensitivity (MedNeXt-L-k5: $0.84 \pm 0.08$; nnU-Net: $0.85 \pm 0.08$) and precision (MedNeXt-L-k5: $0.92 \pm 0.06$; nnU-Net: $0.93 \pm 0.05$) at the voxel level. Correlations between predicted and manual segmentation counts were exceptionally high for both models (voxel-level $r = 0.99$), indicating close agreement with ground-truth annotations. The small differences in all performance metrics between MedNeXt-L-k5 and nnU-Net were negligible.

Paired Wilcoxon signed-rank testing (Table S12) showed that nnU-Net achieved statistically higher scores than MedNeXt-L-k5 across all voxel- and cluster-level metrics ($p_{FDR} \leq 0.041$). While effect sizes were large for most metrics ($r = 0.61$–$0.95$), the absolute median differences were small ($\leq 0.017$), suggesting the advantage, although consistent across subjects, was of limited practical magnitude.

In contrast, when trained on T1w images, MedNeXt-L-k5 showed substantially reduced performance, with voxel-level Dice dropping to $0.58 \pm 0.09$ and sensitivity and precision also lower compared to the T2w models.

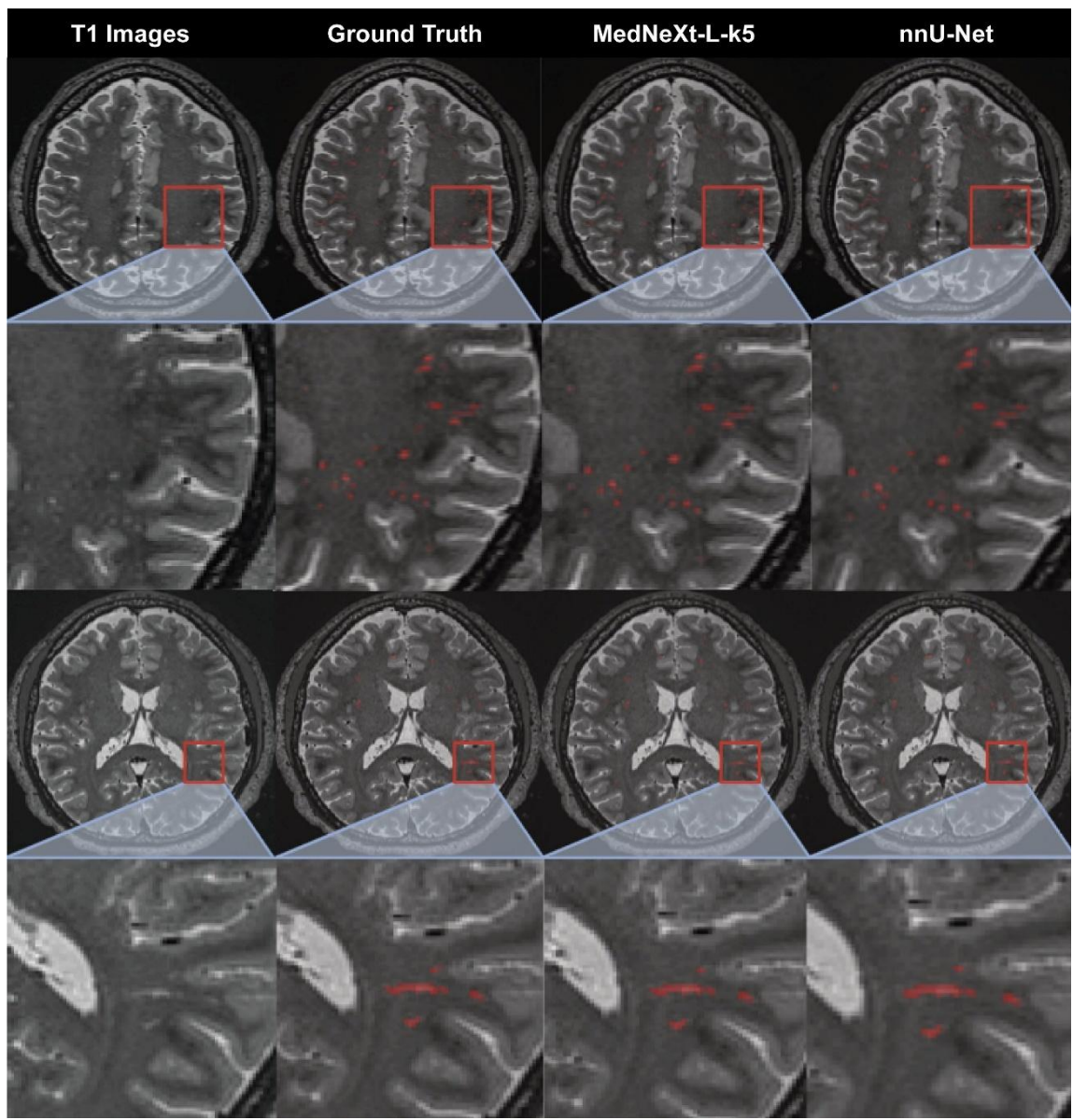

**Fig. 1**: *Overlay of MedNeXt-L-k5 and nnU-Net white matter PVS segmentations on the raw T2 HCP-Aging images with comparison to the ground truth segmentations.*

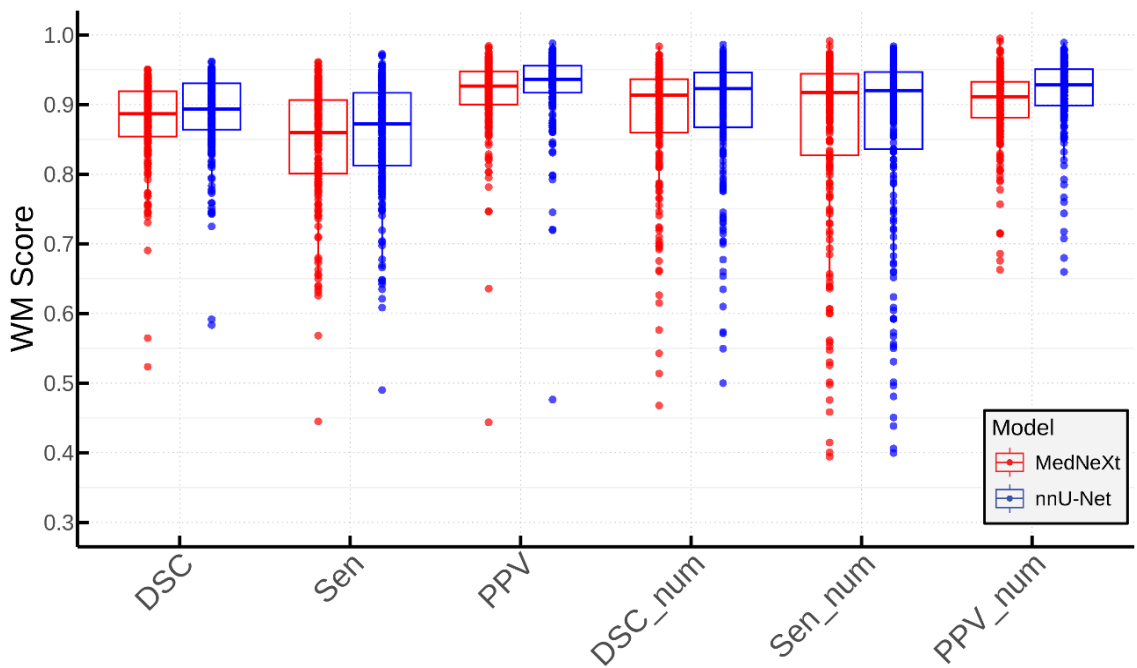

***Fig. 2***: *Performance metrics from internal 5-fold cross-validation (5FCV) in the white matter (WM), on T2-weighted MR images from the HCP-Aging dataset (n = 200), comparing MedNeXt-L-k5 (red) and nnU-Net (blue) segmentation performance. Metrics without the "_num" suffix (DSC, Sen, PPV) are voxel-level measures, whereas metrics with the "_num" suffix (DSC_num, Sen_num, PPV_num) are cluster-level measures. DSC, Dice score; Sen, sensitivity; PPV, positive predictive value; num, cluster-level metrics.*

### 3.2 Internal Validation on In-House T1w Dataset

Table 2 summarizes the internal validation performance metrics of both nnU-Net and MedNeXt-L-k5 trained on our in-house T1w dataset across WM and BG regions. Overall, both models provided robust PVS segmentation (Figure 3), with MedNeXt-L-k5, performing at least on par with nnU-Net and demonstrating slightly higher sensitivity in both WM and BG, although this was accompanied by a modest reduction in precision (Figures S3 and S4).

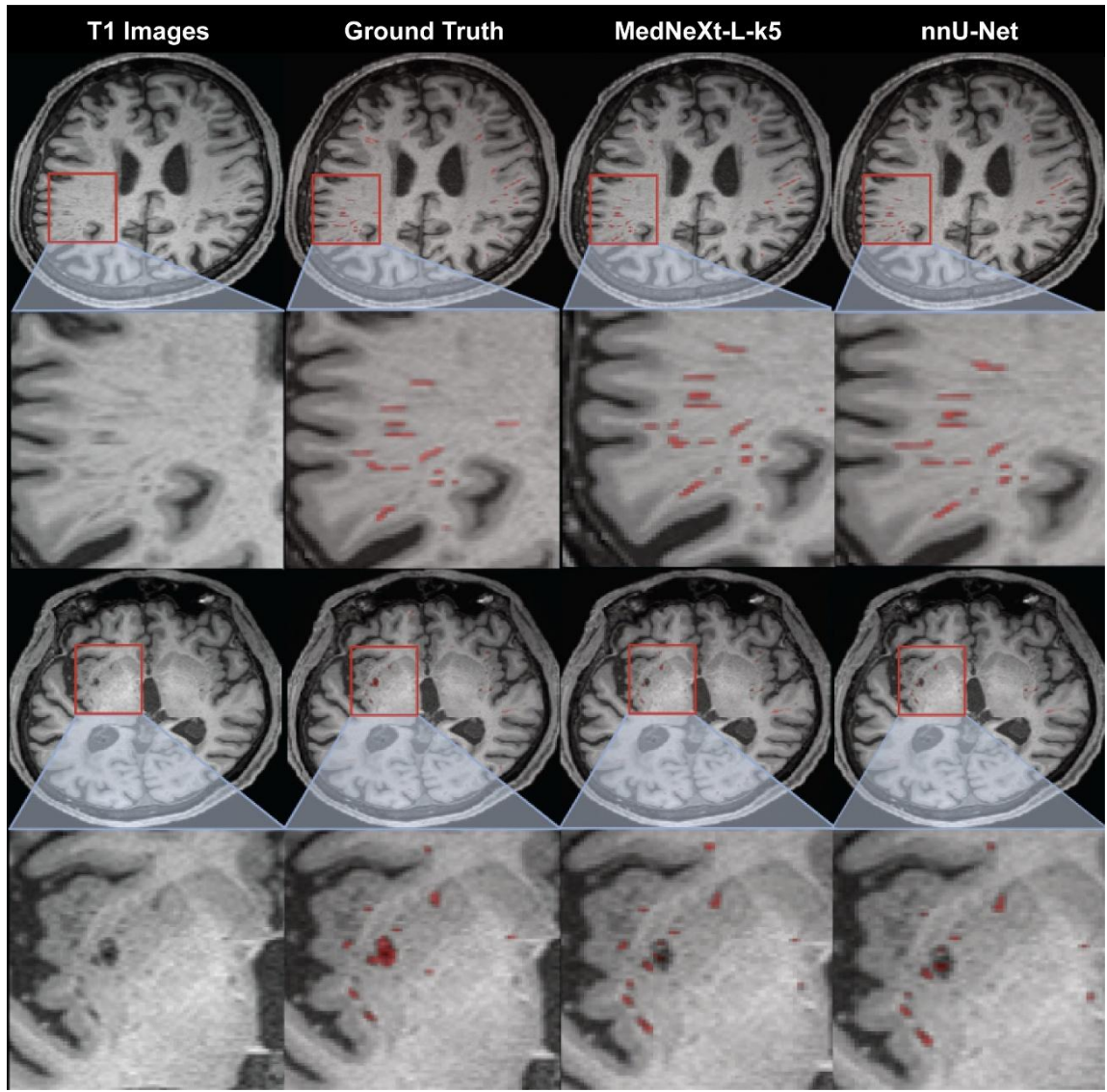

***Fig. 3**: Overlay of MedNeXt-L-k5 and nnU-Net segmentations on the raw T1 images with comparison to the ground truth segmentations.*

In the WM, MedNeXt-L-k5 achieved an average voxel-level Dice score of $0.51 \pm 0.14$ vs. $0.49 \pm 0.14$ for nnU-Net ($p_{FDR} = 0.005$, r = 0.55), with a similar trend at the cluster level ($0.64 \pm 0.17$ vs. $0.63 \pm 0.18$, $p_{FDR}$ 0.413, ns). Sensitivity, the model's ability to detect true PVS voxels, was modestly higher for MedNeXt-L-k5 ($0.43 \pm 0.17$ vs. $0.40 \pm 0.17$, $p_{FDR} < 0.001$, r = 0.78), whereas precision was slightly higher for nnU-Net ($0.74 \pm 0.13$ vs. $0.71 \pm 0.14$, $p_{FDR} = 0.005$, r = -0.57). Correlations between predicted and manual counts were highly similar between models (see Table 2 for complete results).

A similar pattern was observed for the BG region. Both models achieved nearly identical voxel-level Dice scores ($0.53 \pm 0.11$ vs. $0.52 \pm 0.11$, $p_{FDR} = 0.309$, ns), while MedNeXt-L-k5 again showed higher sensitivity ($0.46 \pm 0.14$ vs. $0.43 \pm 0.14$, $p_{FDR} = 0.309$, ns) and nnU-Net maintained higher precision ($0.72 \pm 0.12$ vs. $0.74 \pm 0.10$, $p_{FDR} = 0.035$, r = -0.41).

*Table 2: Internal validation performance metrics for MedNeXt-L configured with a 5x5x5 kernel (MedNeXt-L-k5) and for nnU-Net across different training and evaluation schedules. Metrics are reported as the mean (standard deviation), where the mean is calculated over subjects for Dice Score, Sensitivity, and Precision, and over subjects for correlation.*

| Region | Algorithm | Training Dataset | Evaluation Dataset | Dice Score | | Sensitivity | | Precision | | Correlation | |
|---|---|---|---|---|---|---|---|---|---|---|---|
| | | | | Voxel | Number | Voxel | Number | Voxel | Number | Voxel | Number |
| White Matter | nnU-Net | All Sites | ADNI | 0.44(0.13) | 0.71(0.12) | 0.32(0.12) | 0.60(0.15) | 0.79(0.09) | 0.80(0.09) | - | - |
| | | | AF | 0.60(0.09) | 0.68(0.11) | 0.55(0.15) | 0.68(0.14) | 0.74(0.13) | 0.70(0.19) | - | - |
| | | | ASC | 0.56(0.09) | 0.80(0.08) | 0.43(0.11) | 0.68(0.13) | 0.86(0.05) | 0.85(0.10) | - | - |
| | | | CGS | 0.34(0.16) | 0.34(0.16) | 0.24(0.13) | 0.23(0.12) | 0.69(0.16) | 0.69(0.19) | - | - |
| | | | FTD | 0.35(0.09) | 0.47(0.14) | 0.23(0.07) | 0.33(0.13) | 0.75(0.08) | 0.86(0.06) | - | - |
| | | | MCIS | 0.54(0.08) | 0.67(0.09) | 0.50(0.09) | 0.63(0.10) | 0.61(0.11) | 0.71(0.14) | - | - |
| | | | All Sites | 0.49(0.15) | 0.63(0.18) | 0.40(0.17) | 0.57(0.21) | **0.74(0.14)** | **0.76(0.15)** | 0.81 | 0.76 |
| | MedNeXt-L-k5 | All Sites | ADNI | 0.51(0.11) | 0.79(0.09) | 0.39(0.12) | 0.70(0.08) | 0.78(0.08) | 0.79(0.08) | - | - |
| | | | AF | 0.62(0.04) | 0.69(0.08) | 0.55(0.06) | 0.65(0.11) | 0.74(0.14) | 0.68(0.20) | - | - |
| | | | ASC | 0.49(0.07) | 0.73(0.09) | 0.40(0.10) | 0.65(0.08) | 0.84(0.04) | 0.82(0.06) | - | - |
| | | | CGS | 0.43(0.12) | 0.43(0.16) | 0.31(0.14) | 0.30(0.13) | 0.69(0.14) | 0.66(0.16) | - | - |
| | | | FTD | 0.45(0.10) | 0.47(0.12) | 0.33(0.12) | 0.33(0.13) | 0.73(0.08) | 0.78(0.08) | - | - |
| | | | MCIS | 0.58(0.07) | 0.62(0.08) | 0.60(0.09) | 0.64(0.13) | 0.52(0.12) | 0.67(0.21) | - | - |
| | | | All Sites | **0.51(0.14)** | **0.64(0.18)** | **0.44(0.18)** | **0.60(0.21)** | 0.72(0.16) | 0.73(0.15) | 0.80 | 0.76 |
| Basal Ganglia | nnU-Net | All Sites | ADNI | 0.50(0.09) | 0.77(0.10) | 0.38(0.11) | 0.66(0.09) | 0.78(0.09) | 0.75(0.13) | - | - |
| | | | AF | 0.62(0.04) | 0.67(0.11) | 0.56(0.06) | 0.64(0.11) | 0.71(0.07) | 0.63(0.12) | - | - |
| | | | ASC | 0.44(0.12) | 0.76(0.12) | 0.33(0.13) | 0.66(0.12) | 0.77(0.11) | 0.72(0.17) | - | - |
| | | | CGS | 0.40(0.11) | 0.41(0.19) | 0.28(0.12) | 0.29(0.16) | 0.82(0.09) | 0.79(0.16) | - | - |
| | | | FTD | 0.44(0.06) | 0.44(0.12) | 0.31(0.07) | 0.31(0.12) | 0.81(0.09) | 0.79(0.11) | - | - |
| | | | MCIS | 0.58(0.07) | 0.65(0.07) | 0.54(0.09) | 0.62(0.12) | 0.64(0.09) | 0.74(0.12) | - | - |
| | | | All Sites | 0.52(0.11) | 0.64(0.17) | 0.43(0.14) | 0.56(0.18) | **0.74(0.10)** | 0.72(0.14) | 0.88 | 0.66 |
| | MedNeXt-L-k5 | All Sites | ADNI | 0.46(0.12) | 0.73(0.09) | 0.34(0.12) | 0.64(0.13) | 0.76(0.09) | 0.76(0.12) | - | - |
| | | | AF | 0.61(0.09) | 0.69(0.12) | 0.56(0.15) | 0.71(0.14) | 0.72(0.07) | 0.65(0.11) | - | - |
| | | | ASC | 0.57(0.09) | 0.80(0.07) | 0.44(0.11) | 0.69(0.11) | 0.69(0.12) | 0.69(0.09) | - | - |
| | | | CGS | 0.35(0.18) | 0.35(0.15) | 0.28(0.17) | 0.24(0.12) | 0.79(0.08) | 0.84(0.16) | - | - |

| | | | | | | | | |
|---|---|---|---|---|---|---|---|---|
| FTD | 0.38(0.10) | 0.48(0.16) | 0.26(0.10) | 0.35(0.15) | 0.81(0.13) | 0.83(0.14) | - | - |
| MCIS | 0.53(0.09) | 0.66(0.10) | 0.58(0.07) | 0.70(0.09) | 0.56(0.10) | 0.71(0.11) | - | - |
| All Sites | **0.53(0.11)** | **0.65(0.16)** | **0.46(0.14)** | **0.58(0.18)** | 0.72(0.12) | **0.73(0.13)** | 0.88 | 0.67 |

*3.3 External Validation on In-House T1w Dataset*

In the WM region, the externally trained algorithms SHIVA, WPSS, mc-PVSnet, and MedNeXt-L-k5 trained on HCP-A achieved substantially lower voxel-level Dice scores, ($0.18 \pm 0.13$, $0.30 \pm 0.10$, $0.33 \pm 0.14$, and $0.35 \pm 0.14$, respectively) compared to the LOSOCV-trained nnU-Net ($0.38 \pm 0.15$) and MedNeXt-L-k5 (Dice: $0.38 \pm 0.16$). At the cluster level, both nnU-Net and MedNeXt-L-k5 achieved substantially higher Dice scores ($0.60 \pm 0.19$ and $0.61 \pm 0.19$ respectively) than all external models (Table 3). Voxel-level sensitivity and precision were similar between nnU-Net and MedNeXt-L-k5 (~0.34 sensitivity, ~0.71 precision), and volumetric correlation with manual segmentations was high for both (0.91 for MedNeXt-L-k5 vs. 0.90 for nnU-Net).

Because mc-PVSnet's voxel-level Dice ($0.33 \pm 0.14$) was closer to the LOSOCV models than SHIVA or WPSS, we formally compared it to both nnU-Net and MedNeXt-L-k5 using paired Wilcoxon signed-rank tests (Table S13). In WM, both LOSOCV models achieved significantly higher voxel- and cluster-level Dice and sensitivity scores than mcPVSNet ($p_{FDR} \leq 0.041$). Precision differences were generally non-significant.

In the BG region, the performance gap between mc-PVSNet and the LOSOCV models was larger, with significantly higher voxel- and cluster-level Dice and sensitivity for nnU-Net and MedNeXt-L-k5 ($p_{FDR} \leq 0.043$). As in WM, precision differences were small and often non-significant.

For SHIVA and WPSS, whose Dice scores were far lower than any LOSOCV model, statistical testing was not performed, as differences were visually and numerically large across all metrics.

*3.4 Pre-processing*

Pre-processing (brain extraction, bias field correction, intensity normalisation) slightly reduced segmentation performance (see Table S4), with the model trained and evaluated on preprocessed T1w images achieving a lower mean Dice score ($0.49 \pm 0.14$) compared to the model trained and evaluated on raw T1w images ($0.51 \pm 0.14$).

*Table 3:* *External validation performance metrics for MedNeXt-L configured with a 5x5x5 kernel (MedNeXt-L-k5), nnU-Net, SHIVA, WPSS, and mc-PVSnet across different training and evaluation schedules. Each metric is reported as the mean (standard deviation), where the mean is calculated over subjects for Dice Score, Sensitivity, and Precision, and correlation. The best-performing algorithms are highlighted in bold.* MedNeXt-L-k5 (trained on HCP) does not give BG PVS segmentations.*

| Region | Algorithm | Training Dataset | Evaluation Dataset | Dice Score | | Sensitivity | | Precision | | Correlation | |
|---|---|---|---|---|---|---|---|---|---|---|---|
| | | | | Voxel | Number | Voxel | Number | Voxel | Number | Voxel | Number |
| White Matter | SHIVA | External | All sites | 0.18(0.13) | 0.29(0.19) | 0.12(0.10) | 0.21(0.16) | 0.67(0.19) | 0.77(0.20) | 0.80 | 0.75 |
| | WPSS | External | All sites | 0.30(0.10) | 0.38(0.18) | 0.27(0.10) | 0.68(0.21) | 0.40(0.17) | 0.28(0.17) | 0.86 | 0.49 |
| | mc-PVSnet | External | All sites | 0.33(0.14) | 0.50(0.17) | 0.24(0.13) | 0.36(0.16) | 0.72(0.20) | 0.81(0.17) | 0.83 | 0.82 |
| | MedNeXt-L-k5 (trained on HCP) | External | All sites | 0.35(0.14) | 0.54(0.17) | 0.30(0.16) | 0.49(0.19) | 0.56(0.17) | 0.65(0.17) | 0.91 | 0.85 |
| | nnU-Net (PINGU) | Remaining 5 sites | LOSOCV x 6 | 0.38(0.15) | 0.60(0.19) | 0.34(0.21) | 0.53(0.21) | 0.71(0.25) | 0.75(0.19) | 0.90 | 0.85 |
| | MedNeXt-L-k5 | Remaining 5 sites | LOSOCV x 6 | 0.38(0.16) | 0.61(0.19) | 0.34(0.22) | 0.54(0.20) | 0.71(0.25) | 0.74(0.18) | 0.91 | 0.85 |
| Basal Ganglia | SHIVA | External | All sites | 0.10(0.11) | 0.15(0.09) | 0.07(0.09) | 0.09(0.05) | 0.61(0.28) | 0.79(0.30) | 0.11 | 0.00 |
| | WPSS | External | All sites | 0.20(0.09) | 0.37(0.17) | 0.15(0.10) | 0.56(0.19) | 0.39(0.17) | 0.28(0.15) | 0.56 | 0.34 |
| | mc-PVSnet | External | All sites | 0.28(0.13) | 0.54(0.19) | 0.20(0.12) | 0.37(0.15) | 0.74(0.17) | 0.85(0.14) | 0.68 | 0.69 |
| | MedNeXt-L-k5 (trained on HCP) * | - | - | - | - | - | - | - | - | - | - |
| | nnU-Net (PINGU) | Remaining 5 sites | LOSOCV x 6 | 0.35(0.12) | 0.60(0.20) | 0.34(0.24) | 0.50(0.18) | 0.70(0.26) | 0.75(0.19) | 0.93 | 0.78 |

| MedNeXt-L-k5 | Remaining 5 sites | LOSOCV x 6 | 0.35(0.12) | 0.62(0.21) | 0.34(0.25) | 0.52(0.18) | 0.70(0.26) | 0.76(0.20) | 0.94 | 0.79 |

## 4. DISCUSSION

The development of robust, automated methods for perivascular space (PVS) segmentation is increasingly important in neuroimaging, given the growing evidence linking PVS burden to cerebral small vessel disease and neurodegeneration. Here, we present the first deep learning PVS segmentation model trained on a T2-weighted dataset of this scale ($n$ = 200), the largest to date, and demonstrate state-of-the-art segmentation performance (mean Dice = 0.88), substantially exceeding previous reports in the literature.

Our MedNeXt-L-k5 model, based on a large-kernel ConvNeXt segmentation architecture, not only achieved these high Dice scores on T2w images, but also showed robust and competitive generalization on heterogeneous T1w datasets (mean Dice = 0.51±0.14), comparable to nnU-Net (mean Dice = 0.49±0.14), and outperformed alternatives such as SHIVA, WPSS and mcPVS-Net in both internal and external validation.

A key strength of this study is the unprecedented scale of T2-weighted (T2w) images used for model development. With 200 T2w images, this represents the largest dataset yet assembled for deep learning-based PVS segmentation, yielding higher Dice scores than that reported in the literature for T2w PVS segmentation with deep learning. For example, Cai et al. (2024) reported a Dice score of 0.702, while Choi et al. (2020) did not report a Dice score, instead presenting within-pair intraclass correlation coefficients (ICCs) for deep PVS volumes that reflect heritability rather than segmentation accuracy. MedNeXt-L-k5 achieved a high mean Dice score on T2w images (mean Dice 0.88), far higher than performance on T1w images both on the HCP-A T1w images (mean Dice 0.58) and our in-house T1w dataset (mean Dice = 0.51). Several factors likely contribute to this difference: (i) greater anatomical diversity in the in-house T1w cohort; (ii) the inclusion of only healthy controls and exclusion of large lacunes in the T2w dataset; (iii) the inherently higher contrast of PVS in T2w images, which facilitates clearer boundary definition compared to T1w scans, where PVS are less distinguishable from WMH or small vessels; (iv) the manual annotations in the HCP-A dataset were semi-automated, it may be that semi-automated annotations are easier to reproduce than fully manual annotations, and (v) the ground truth segmentations on the HCP-A dataset were annotated on the T2w images, and their transferal to the T1w images may introduce interpolation errors. It is also important to note that PVS segmentation in the HCP-Aging T2w dataset was limited to white matter, while our T1w analysis included both WM and BG, adding further complexity to the latter.

Our findings also highlight the nuanced role of pre-processing in medical image segmentation. While pre-processing can reduce noise and standardize intensity distributions (Raad et al., 2021), in our case we found that pre-processing conferred no benefit, (Dice score decreased from $0.51\pm0.14$ to $0.49\pm0.14$). This suggests that MedNeXt-L-k5 is robust to variations in image quality, potentially due to its 5x5x5 kernel architecture, which provides an expanded receptive field. By aggregating information over a large area, the model may be less affected by local noise or artifacts, enabling it to better capture the tubular morphology of PVS. This result is consistent with recent studies reporting that some deep learning architectures, particularly those using deep convolutional neural networks and vision transformers, receive negligible or even detrimental effects in segmentation performance from pre-processing (Kondrateva et al., 2024). For small, low-contrast structures like PVS, there is a risk that aggressive pre-processing may obscure essential details. It is also possible that the preprocessing steps applied, such as Fuzzy C-means intensity normalization and bias field correction, failed to enhance PVS visibility or inadvertently introduced artifacts that negated potential benefits (Sepehrband et al., 2019).

Nonetheless, this study has several limitations. While our T2w model was trained on a relatively large sample of high-resolution, expert-refined annotations ($n = 200$), segmentation was limited exclusively to PVS in the WM. PVS in deep gray matter regions, including the basal ganglia and hippocampus, were not annotated, which restricts the applicability of models trained on this dataset for whole-brain PVS analysis or investigations into region-specific disease processes. This focus reflects a broader challenge in the field, as accurate PVS annotation outside white matter remains time-consuming and ambiguous. Additionally, the HCP-Aging dataset limited by its single-site, single-scanner design and the healthy-aging bias of its population. All subjects were scanned on a Siemens 3T Prisma at a single centre and were selected to exclude individuals with major neurological or psychiatric disease, symptomatic stroke, or dementia. As a result, the cohort under-represents the full spectrum of pathologies typically encountered in clinical neuroimaging, especially those with high PVS burden or severe small vessel disease. While this homogeneity promotes consistency in imaging quality and annotation, it may restrict the generalizability of models trained on this dataset to other scanners, acquisition protocols, or more clinically diverse populations.

In contrast, our T1w PVS segmentation models were developed using a clinically diverse sample collected across seven different scanners from six independent sites, with the aim of achieving greater generalizability across imaging conditions and patient populations. Unlike many previous studies that have focused solely on cognitively healthy populations, our T1w in-house dataset includes a broader clinical spectrum, encompassing healthy controls, individuals with mild cognitive impairment (MCI), and various dementia subtypes. This diversity increases both the challenge and clinical relevance of the segmentation task, as accurate PVS quantification is most crucial in patients with neurological disease, where PVS must be distinguished from imaging mimics such as lacunes and white matter hyperintensities (WMH). Furthermore, our model was trained on scans from seven scanners across six sites, specifically to maximize generalizability. The consistent performance observed across these heterogeneous imaging conditions further supports MedNeXt-L-k5's robustness for real-world applications.

However, these models also have their limitations. Chief among these is the limited size of the annotated T1w dataset ($n = 40$). Manual annotation of PVS is time-consuming and subject to ambiguity due to their small size, high frequency, and visual similarity to other lesions. Each T1w scan comprises hundreds of slices across all three anatomical planes, requiring detailed and time-consuming review by expert raters. This limitation is common to most current PVS segmentation studies, including mc-PVSnet (Huang et al., 2024), SHIVA-PVS (Boutinaud et al., 2021), and PINGU (Sinclair et al., 2024), which are also trained on datasets of only 40-50 scans. Our use of a more heterogeneous sample is intended to increase the external generalizability of our models. However, greater diversity can also introduce additional label variance, which may modestly reduce the statistical power.

Another important limitation is the lack of explicit of inter- and intra-rater reliability assessment in the T1w dataset. While inter-rater reliability was high for the T2w dataset (Cohen's κ = 0.95±0.02; Chai et al. (2025)), the T1w annotations were contributed by a relatively large pool of raters, with insufficient overlap to reliably assess annotation consistency across different raters or within individual raters over time. Previous studies have reported moderate to excellent reliability for T1w PVS annotations, with intra-rater agreement (Cohen's κ) approaching 0.80 in manual segmentation protocols (Dubost, Adams, et al., 2019) and inter-scan intraclass correlation coefficients (ICC) as high as 0.966 for volumetric measures (Huang et al., 2024). This omission leaves open the possibility of label variability, which may impact both model training and evaluation. On one hand, involving many raters may capture diverse expert interpretations, potentially distilling common PVS features and reducing individual bias. On the other hand, it risks introducing inconsistency, especially with the challenge of distinguishing PVS from mimics like WMH, lacunes, and microbleeds on T1w images. Raters were trained to differentiate these structures, but the lack of complementary T2w or FLAIR images, which better resolve such mimics, likely increased annotation difficulty and variability.

Scanner diversity was also high in our study, encompassing a range of vendors and protocols. While this heterogeneity may be beneficial for encouraging model robustness and generalization to real-world data, the dataset size may not be large enough to perform meaningful subgroup analysis by scanner model or vendor. We did not train separate models for individual sites to compare segmentation performance, as some cohorts such as ASC, MCIS, and FTD had fewer than five samples for cross-validation. This also prevented us from assessing correlations between segmented and actual PVS volumes in some cohorts due to insufficient statistical power.

The MedNeXt-L-k5 architecture, with its larger 5x5x5 kernels and attention mechanisms, was initially hypothesized to better capture the elongated, tubular structure of PVS across WM and BG regions. In practice, however, our results show that MedNeXt-L-k5 and nnU-Net achieve nearly identical performance across all validation sites. In WM segmentation, for example, the mean Dice scores were 0.51±0.14 for MedNeXt-L-k5 and 0.49±0.14 for nnU-Net (Table 2), a difference well within the margin of error. Other performance metrics, including sensitivity, specificity, and correlation, were likewise comparable: MedNeXt-L-k5 demonstrated marginally higher sensitivity, while nnU-Net was slightly more specific at both voxel and cluster levels.

The strong correlations between MedNeXt-L-k5-derived PVS volumes/clusters and manual segmentations in both T2w (r = 0.99 for WM) and T1w (r = 0.90-0.95 for WM and r = 0.85-0.91 for BG) images (Tables 2 and S3) support its use as a surrogate for manual ratings in research and clinical studies. This is particularly valuable given the limitations of traditional visual rating scales, which are subject to inter-rater variability and offer only coarse estimates of PVS burden rather than precise segmentation maps with exact volumetric measurements. However, it is worth noting that MedNeXt-L-k5, as the most computationally intensive subvariant of the MedNeXt family, delivers superior accuracy at the cost of increased resource demands. While well-suited for research settings with robust computational infrastructure, its practical deployment in some clinical environments may be constrained compared to lighter subvariants, such as MedNeXt-S or MedNeXt-M, as well as on smaller kernels (e.g., 3x3x3), which may offer a more feasible balance of accuracy and computational efficiency.

The large receptive field and long-range context inherent to the MedNeXt-L give the model a greater capacity to capture global and contextual features, such as anatomical location, image overall contrast to noise etc. Given that PVS appearance and detection are sensitive to such features, we expected this to result in improved performance compared to its industry-leading peer, the nnU-Net, but this was not the case; the performance scores were very similar across all metrics and experiments. We interpret this to reflect the strength of the nnU-Net, which yields field-leading performances across a range of tasks (Isensee et al., 2021). It may be that both algorithms have hit the ceiling of the maximum possible performance given the datasets. In particular, the inter-rater reliability of the manual PVS segmentations used as the ground truth places a limit on the performance of a trained network. Alternatively, it may be that different PVS are similar enough in appearance across anatomical locations and global signal to noise etc., that taking these into account is not required to achieve a good segmentation.

Nonetheless, the high Dice scores achieved, especially on T2w images (0.88), highlight MedNeXt-L-k5's strong potential for high-throughput, automated screening of PVS burden in population studies. Since PVS enlargement has been associated with an increased risk of neurodegenerative conditions such as cerebral small vessel disease and Alzheimer's disease (Lynch et al., 2022), accurate quantification of PVS volume may aid early detection and monitoring of neurovascular pathology, thereby informing preventative strategies.

## 5. CONCLUDING REMARKS

In summary, MedNeXt-L-k5 demonstrates robust performance on both our heterogeneous internal T1w dataset ($n = 40$) and the T2w HCP-Aging dataset ($n = 200$), achieving Dice, sensitivity, precision, and correlation scores that match or exceed current leading segmentation frameworks. Its strong generalizability across multi-site datasets, despite substantial clinical and imaging heterogeneity, positions it as a scalable, reliable tool for PVS quantification in large neuroimaging studies. Our T1w model generalizes well across diverse populations, while our T2w model achieves state-of-the-art accuracy (Dice = 0.88) for white matter PVS segmentation. Future research should explore formal benchmarking against human raters, validation in larger and more diverse clinical cohorts, expansion from white matter-only to whole-brain PVS segmentation, and integration into real-world research and clinical workflows.

**Data and Code Availability**

The HCP-A, T2w and T1w models, named MedNet-PVS (MedNeXt-based network for automated perivascular space segmentation), are available at https://github.com/iBrain-Lab/MedNet-PVS, with pretrained weights downloadable at doi: 10.26180/29588624.

**CRediT authorship contribution statement**

**Zhen Xuen Brandon Low:** Conceptualization, Methodology, Software, Formal analysis, Investigation, Data Curation, Visualization, Writing - original draft, Writing – review & editing. **Rory Zhang:** Software, Investigation, Writing – review & editing. **Hang Min, Ying Xia:** Resources, Methodology, Writing – review & editing. **William Pham, Lucy Vivash, Jasmine Moses, Miranda Lynch, Karina Dorfman, Cassandra Marotta, Shaun Koh, Jacob Bunyamin, Ella Rowsthorn, Alex Jarema, Himashi Peiris, Zhaolin Chen, Sandy R Shultz, David K Wright, Dexiao Kong, Sharon L. Naismith, Terence J. O'Brien:** Data Curation, Writing – review & editing. **Meng Law**: Supervision, Resources, Investigation, Validation, Writing – review & editing. **Benjamin Sinclair:** Conceptualization, Methodology, Software, Validation, Resources, Supervision, Project administration, Writing – review & editing.

**Declaration of competing interest**

The authors declare that there are no known competing financial interests or personal relationships that could have appeared to influence the work reported in this paper.

## REFERENCES

Aribisala, B. S., Valdés Hernández, M. d. C., Okely, J. A., Cox, S. R., Ballerini, L., Dickie, D. A., Wiseman, S. J., Riha, R. L., Muñoz Maniega, S., Radakovic, R., Taylor, A., Pattie, A., Corley, J., Redmond, P., Bastin, M. E., Deary, I., & Wardlaw, J. M. (2023). Sleep quality, perivascular spaces and brain health markers in ageing - A longitudinal study in the Lothian Birth Cohort 1936. *Sleep Medicine*, *106*, 123-131. https://doi.org/https://doi.org/10.1016/j.sleep.2023.03.016

Avants, B. B., Epstein, C. L., Grossman, M., & Gee, J. C. (2008). Symmetric diffeomorphic image registration with cross-correlation: Evaluating automated labeling of elderly and neurodegenerative brain. *Medical Image Analysis*, *12*(1), 26-41. https://doi.org/https://doi.org/10.1016/j.media.2007.06.004

Bookheimer, S. Y., Salat, D. H., Terpstra, M., Ances, B. M., Barch, D. M., Buckner, R. L., Burgess, G. C., Curtiss, S. W., Diaz-Santos, M., Elam, J. S., Fischl, B., Greve, D. N., Hagy, H. A., Harms, M. P., Hatch, O. M., Hedden, T., Hodge, C., Japardi, K. C., Kuhn, T. P., . . . Yacoub, E. (2019). The Lifespan Human Connectome Project in Aging: An overview. *Neuroimage*, *185*, 335-348. https://doi.org/10.1016/j.neuroimage.2018.10.009

Boutinaud, P., Tsuchida, A., Laurent, A., Adonias, F., Hanifehlou, Z., Nozais, V., Verrecchia, V., Lampe, L., Zhang, J., Zhu, Y.-C., Tzourio, C., Mazoyer, B., & Joliot, M. (2021). 3D Segmentation of Perivascular Spaces on T1-Weighted 3 Tesla MR Images With a Convolutional Autoencoder and a U-Shaped Neural Network [Original Research]. *Frontiers in Neuroinformatics*, *15*. https://doi.org/10.3389/fninf.2021.641600

Cai, D., Pan, M., Liu, C., He, W., Ge, X., Lin, J., Li, R., Liu, M., & Xia, J. (2024). Deep-learning-based segmentation of perivascular spaces on T2-Weighted 3T magnetic resonance images. *Front Aging Neurosci*, *16*, 1457405. https://doi.org/10.3389/fnagi.2024.1457405

Chae, S., Yang, E., Moon, W. J., & Kim, J. H. (2024). Deep Cascade of Convolutional Neural Networks for Quantification of Enlarged Perivascular Spaces in the Basal Ganglia in Magnetic Resonance Imaging. *Diagnostics*, *14*(14), Article 1504. https://doi.org/10.3390/diagnostics14141504

Chai, Y., Zhang, H., Robles, C., Kim, A. S., Janhanshad, N., Thompson, P. M., van der Werf, Y., van Heese, E. M., Kim, J., Joo, E. Y., Aksman, L., Kang, K.-W., Shin, J.-W., Trang, A., Ha, J., Lee, E., Moon, Y., & Kim, H. (2025). Precise perivascular space segmentation on magnetic resonance imaging from Human Connectome Project-Aging. *medRxiv*, 2025.2003.2019.25324269. https://doi.org/10.1101/2025.03.19.25324269

Choi, Y., Nam, Y., Choi, Y., Kim, J., Jang, J., Ahn, K. J., Kim, B.-s., & Shin, N.-Y. (2020). MRI-visible dilated perivascular spaces in healthy young adults: A twin heritability study. *Human Brain Mapping*, *41*(18), 5313-5324. https://doi.org/https://doi.org/10.1002/hbm.25194

Çiçek, Ö., Abdulkadir, A., Lienkamp, S. S., Brox, T., & Ronneberger, O. (2016). 3D U-Net: learning dense volumetric segmentation from sparse annotation. International conference on medical image computing and computer-assisted intervention,

Desikan, R. S., Ségonne, F., Fischl, B., Quinn, B. T., Dickerson, B. C., Blacker, D., Buckner, R. L., Dale, A. M., Maguire, R. P., Hyman, B. T., Albert, M. S., & Killiany, R. J. (2006). An automated labeling system for subdividing the human cerebral cortex on MRI scans into gyral based regions of interest. *Neuroimage*, *31*(3), 968-980. https://doi.org/https://doi.org/10.1016/j.neuroimage.2006.01.021

Duarte Coello, R., Valdés Hernández, M. d. C., Zwanenburg, J. J. M., van der Velden, M., Kuijf, H. J., De Luca, A., Moyano, J. B., Ballerini, L., Chappell, F. M., Brown, R., Jan Biessels, G., & Wardlaw, J. M. (2024). Detectability and accuracy of computational measurements of in-silico and physical representations of enlarged perivascular spaces from magnetic resonance images. *Journal of Neuroscience Methods*, *403*, 110039. https://doi.org/https://doi.org/10.1016/j.jneumeth.2023.110039

Dubost, F., Adams, H., Bortsova, G., Ikram, M. A., Niessen, W., Vernooij, M., & de Bruijne, M. (2019). 3D regression neural network for the quantification of enlarged perivascular spaces in brain MRI. *Medical Image Analysis*, *51*, 89-100. https://doi.org/https://doi.org/10.1016/j.media.2018.10.008

Dubost, F., Adams, H., Yilmaz, P., Bortsova, G., Tulder, G. v., Ikram, M. A., Niessen, W., Vernooij, M. W., & Bruijne, M. d. (2020). Weakly supervised object detection with 2D and 3D regression neural networks. *Medical Image Analysis*, *65*, 101767. https://doi.org/https://doi.org/10.1016/j.media.2020.101767

Dubost, F., Bortsova, G., Adams, H., Ikram, A., Niessen, W. J., Vernooij, M., & De Bruijne, M. (2017). Gp-unet: Lesion detection from weak labels with a 3d regression network. International Conference on Medical Image Computing and Computer-Assisted Intervention,

Dubost, F., Dünnwald, B., Huff, D., Scheumann, V., Schreiber, F., Vernooij, M., Niessen, W., Skalej, M., Schreiber, S., Oeltze-Jafra, S., & de Bruijne, M. (2019). Automated Quantification of Enlarged Perivascular Spaces in Clinical Brain MRI Across Sites. In L. Zhou, D. Sarikaya, S. M. Kia, S. Speidel, A. Malpani, D. Hashimoto, M. Habes, T. Löfstedt, K. Ritter, & H. Wang, *OR 2.0 Context-Aware Operating Theaters and Machine Learning in Clinical Neuroimaging* Cham.

Dubost, F., Yilmaz, P., Adams, H., Bortsova, G., Ikram, M. A., Niessen, W., Vernooij, M., & de Bruijne, M. (2019). Enlarged perivascular spaces in brain MRI: Automated quantification in four regions. *Neuroimage*, *185*, 534-544. https://doi.org/10.1016/j.neuroimage.2018.10.026

Gyanwali, B., Vrooman, H., Venketasubramanian, N., Wong, T. Y., Cheng, C.-Y., Chen, C., & Hilal, S. (2019). Cerebral Small Vessel Disease and Enlarged Perivascular Spaces-Data From Memory Clinic and Population-Based

Settings [Original Research]. *Frontiers in Neurology*, *10*. https://doi.org/10.3389/fneur.2019.00669

He, K., Gan, C., Li, Z., Rekik, I., Yin, Z., Ji, W., Gao, Y., Wang, Q., Zhang, J., & Shen, D. (2023). Transformers in medical image analysis. *Intelligent Medicine*, *3*(1), 59-78. https://doi.org/https://doi.org/10.1016/j.imed.2022.07.002

Hernández, M. D. V., Coello, R. D., Xu, W. L., Bernal, J., Cheng, Y. J., Ballerini, L., Wiseman, S. J., Chappell, F. M., Clancy, U., Garcia, D. J., Reyes, C. A., Zhang, J. F., Liu, X. D., Hewins, W., Stringer, M., Doubal, F., Thrippleton, M. J., Jochems, A., Brown, R., & Wardlaw, J. M. (2024). Influence of threshold selection and image sequence in in-vivo segmentation of enlarged perivascular spaces. *Journal of Neuroscience Methods*, *403*, Article 110037. https://doi.org/10.1016/j.jneumeth.2023.110037

Huang, P. Y., Liu, L. Y., Zhang, Y., Zhong, S. Y., Liu, P., Hong, H., Wang, S. Y., Xie, L. Y., Lin, M., Jiaerken, Y., Luo, X., Li, K. C., Zeng, Q. Z., Cui, L., Li, J. X., Chen, Y. X., Zhang, R. T., & Alzheimers Dis Neuroimaging, I. (2024). Development and validation of a perivascular space segmentation method in multi-center datasets. *Neuroimage*, *298*, Article 120803. https://doi.org/10.1016/j.neuroimage.2024.120803

Isensee, F., Jaeger, P. F., Kohl, S. A. A., Petersen, J., & Maier-Hein, K. H. (2021). nnU-Net: a self-configuring method for deep learning-based biomedical image segmentation. *Nature Methods*, *18*(2), 203-211. https://doi.org/10.1038/s41592-020-01008-z

Jack, C. R., Jr., Bernstein, M. A., Fox, N. C., Thompson, P., Alexander, G., Harvey, D., Borowski, B., Britson, P. J., J, L. W., Ward, C., Dale, A. M., Felmlee, J. P., Gunter, J. L., Hill, D. L., Killiany, R., Schuff, N., Fox-Bosetti, S., Lin, C., Studholme, C., . . . Weiner, M. W. (2008). The Alzheimer's Disease Neuroimaging Initiative (ADNI): MRI methods. *J Magn Reson Imaging*, *27*(4), 685-691. https://doi.org/10.1002/jmri.21049

Javierre-Petit, C., Kontzialis, M., Leurgans, S. E., Bennett, D. A., Schneider, J. A., & Arfanakis, K. (2024). Quantitative assessment of enlarged perivascular spaces via deep-learning in community-based older adults reveals independent associations with vascular neuropathologies, vascular risk factors and cognition. *Brain Communications*, *6*(4), Article fcae252. https://doi.org/10.1093/braincomms/fcae252

Jenkinson, M., Beckmann, C. F., Behrens, T. E., Woolrich, M. W., & Smith, S. M. (2012). FSL. *Neuroimage*, *62*(2), 782-790. https://doi.org/10.1016/j.neuroimage.2011.09.015

Jung, E., Chikontwe, P., Zong, X., Lin, W., Shen, D., & Park, S. H. (2019). Enhancement of Perivascular Spaces Using Densely Connected Deep Convolutional Neural Network. *IEEE Access*, *7*, 18382-18391. https://doi.org/10.1109/access.2019.2896911

Jung, E., Zong, X., Lin, W., Shen, D., & Park, S. H. (2018). Enhancement of Perivascular Spaces Using a Very Deep 3D Dense Network. In I. Rekik, G. Unal, E. Adeli, & S. H. Park, *PRedictive Intelligence in MEdicine* Cham.

Kondrateva, E., Druzhinina, P., Dalechina, A., Zolotova, S., Golanov, A., Shirokikh, B., Belyaev, M., & Kurmukov, A. (2024). Negligible effect of brain MRI data preprocessing for tumor segmentation. *Biomedical Signal Processing and Control*, *96*, 106599. https://doi.org/https://doi.org/10.1016/j.bspc.2024.106599

Kong, S. D. X., Hoyos, C. M., Phillips, C. L., McKinnon, A. C., Lin, P., Duffy, S. L., Mowszowski, L., LaMonica, H. M., Grunstein, R. R., Naismith, S. L., & Gordon, C. J. (2021). Altered heart rate variability during sleep in mild cognitive impairment. *Sleep*, *44*(4), zsaa232. https://doi.org/10.1093/sleep/zsaa232

Kumar, S. S. (2025). Advancements in medical image segmentation: A review of transformer models. *Computers and Electrical Engineering*, *123*, 110099. https://doi.org/https://doi.org/10.1016/j.compeleceng.2025.110099

Lan, H., Lynch, K. M., Custer, R., Shih, N.-C., Sherlock, P., Toga, A. W., Sepehrband, F., & Choupan, J. (2023). Weakly supervised perivascular spaces segmentation with salient guidance of Frangi filter. *Magnetic Resonance in Medicine*, *89*(6), 2419-2431. https://doi.org/https://doi.org/10.1002/mrm.29593

Laveskog, A., Wang, R., Bronge, L., Wahlund, L. O., & Qiu, C. (2018). Perivascular Spaces in Old Age: Assessment, Distribution, and Correlation with White Matter Hyperintensities. *AJNR Am J Neuroradiol*, *39*(1), 70-76. https://doi.org/10.3174/ajnr.A5455

Li, H., Nan, Y., Del Ser, J., & Yang, G. (2024). Large-Kernel Attention for 3D Medical Image Segmentation. *Cognit Comput*, *16*(4), 2063-2077. https://doi.org/10.1007/s12559-023-10126-7

Lian, C., Zhang, J., Liu, M., Zong, X., Hung, S.-C., Lin, W., & Shen, D. (2018). Multi-channel multi-scale fully convolutional network for 3D perivascular spaces segmentation in 7T MR images. *Medical Image Analysis*, *46*, 106-117. https://doi.org/https://doi.org/10.1016/j.media.2018.02.009

Liu, Z., Mao, H., Wu, C.-Y., Feichtenhofer, C., Darrell, T., & Xie, S. (2022). A convnet for the 2020s. Proceedings of the IEEE/CVF conference on computer vision and pattern recognition,

Lynch, M., Pham, W., Sinclair, B., O'Brien, T. J., Law, M., & Vivash, L. (2022). Perivascular spaces as a potential biomarker of Alzheimer's disease. *Front Neurosci*, *16*, 1021131. https://doi.org/10.3389/fnins.2022.1021131

Major, B., Symons, G. F., Sinclair, B., O'Brien, W. T., Costello, D., Wright, D. K., Clough, M., Mutimer, S., Sun, M., Yamakawa, G. R., Brady, R. D., O'Sullivan, M. J., Mychasiuk, R., McDonald, S. J., O'Brien, T. J., Law, M., Kolbe, S., & Shultz, S. R. (2021). White and Gray Matter Abnormalities in Australian Footballers With a History of Sports-Related Concussion: An MRI Study. *Cereb Cortex*, *31*(12), 5331-5338. https://doi.org/10.1093/cercor/bhab161

Mestre, H., Du, T., Sweeney, A. M., Liu, G., Samson, A. J., Peng, W., Mortensen, K. N., Stæger, F. F., Bork, P. A. R., Bashford, L., Toro, E. R., Tithof, J., Kelley, D. H., Thomas, J. H., Hjorth, P. G., Martens, E. A., Mehta, R. I., Solis, O., Blinder,

P., . . . Nedergaard, M. (2020). Cerebrospinal fluid influx drives acute ischemic tissue swelling. *Science*, *367*(6483). https://doi.org/10.1126/science.aax7171

Mori, S., Wakana, S., Nagae-Poetscher, L., & Van Zijl, P. (2006). MRI atlas of human white matter. *American Journal of Neuroradiology*, *27*(6), 1384.

Nooner, K. B., Colcombe, S., Tobe, R., Mennes, M., Benedict, M., Moreno, A., Panek, L., Brown, S., Zavitz, S., Li, Q., Sikka, S., Gutman, D., Bangaru, S., Schlachter, R. T., Kamiel, S., Anwar, A., Hinz, C., Kaplan, M., Rachlin, A., . . . Milham, M. (2012). The NKI-Rockland Sample: A Model for Accelerating the Pace of Discovery Science in Psychiatry [Review]. *Frontiers in Neuroscience*, *6*. https://doi.org/10.3389/fnins.2012.00152

Perosa, V., Oltmer, J., Munting, L. P., Freeze, W. M., Auger, C. A., Scherlek, A. A., van der Kouwe, A. J., Iglesias, J. E., Atzeni, A., Bacskai, B. J., Viswanathan, A., Frosch, M. P., Greenberg, S. M., & van Veluw, S. J. (2022). Perivascular space dilation is associated with vascular amyloid-β accumulation in the overlying cortex. *Acta Neuropathol*, *143*(3), 331-348. https://doi.org/10.1007/s00401-021-02393-1

Potter, G. M., Chappell, F. M., Morris, Z., & Wardlaw, J. M. (2015). Cerebral Perivascular Spaces Visible on Magnetic Resonance Imaging: Development of a Qualitative Rating Scale and its Observer Reliability. *Cerebrovascular Diseases*, *39*(3-4), 224-231. https://doi.org/10.1159/000375153

Raad, K. B. d., Garderen, K. A. v., Smits, M., Voort, S. R. v. d., Incekara, F., Oei, E. H. G., Hirvasniemi, J., Klein, S., & Starmans, M. P. A. (2021, 13-16 April 2021). The Effect of Preprocessing on Convolutional Neural Networks for Medical Image Segmentation. 2021 IEEE 18th International Symposium on Biomedical Imaging (ISBI),

Ranti, D. L., Warburton, A. J., Rutland, J. W., Dullea, J. T., Markowitz, M., Smith, D. A., Kligler, S. Z. K., Rutter, S., Langan, M., Arrighi-Allisan, A., George, I., Verma, G., Murrough, J. W., Delman, B. N., Balchandani, P., & Morris, L. S. (2022). Perivascular spaces as a marker of psychological trauma in depression: A 7-Tesla MRI study. *Brain Behav*, *12*(7), e32598. https://doi.org/10.1002/brb3.2598

Rashid, T., Liu, H., Ware, J. B., Li, K., Romero, J. R., Fadaee, E., Nasrallah, I. M., Hilal, S., Bryan, R. N., Hughes, T. M., Davatzikos, C., Launer, L., Seshadri, S., Heckbert, S. R., & Habes, M. (2023). Deep learning based detection of enlarged perivascular spaces on brain MRI. *Neuroimage: Reports*, *3*(1), 100162. https://doi.org/https://doi.org/10.1016/j.ynirp.2023.100162

Roy, S., Koehler, G., Ulrich, C., Baumgartner, M., Petersen, J., Isensee, F., Jaeger, P. F., & Maier-Hein, K. H. (2023). Mednext: transformer-driven scaling of convnets for medical image segmentation. International Conference on Medical Image Computing and Computer-Assisted Intervention,

Sepehrband, F., Barisano, G., Sheikh-Bahaei, N., Cabeen, R. P., Choupan, J., Law, M., & Toga, A. W. (2019). Image processing approaches to enhance perivascular space visibility and quantification using MRI. *Scientific Reports*, *9*(1), 12351. https://doi.org/10.1038/s41598-019-48910-x

Sinclair, B., Vivash, L., Moses, J., Lynch, M., Pham, W., Dorfman, K., Marotta, C., Koh, S., Bunyamin, J., & Rowsthorn, E. (2024). Perivascular space Identification Nnunet for Generalised Usage (PINGU). *arXiv preprint arXiv:2405.08337*.

Sotgiu, M. A., Lo Jacono, A., Barisano, G., Saderi, L., Cavassa, V., Montella, A., Crivelli, P., Carta, A., & Sotgiu, S. (2023). Brain perivascular spaces and autism: clinical and pathogenic implications from an innovative volumetric MRI study. *Front Neurosci*, *17*, 1205489. https://doi.org/10.3389/fnins.2023.1205489

Spijkerman, J. M., Zwanenburg, J. J. M., Bouvy, W. H., Geerlings, M. I., Biessels, G. J., Hendrikse, J., Luijten, P. R., & Kuijf, H. J. (2022). Automatic quantification of perivascular spaces in T2-weighted images at 7 T MRI. *Cerebral Circulation - Cognition and Behavior*, *3*, 100142. https://doi.org/https://doi.org/10.1016/j.cccb.2022.100142

Sudre, C. H., Anson, B. G., Ingala, S., Lane, C. D., Jimenez, D., Haider, L., Varsavsky, T., Smith, L., Ourselin, S., & Jäger, R. H. (2019). 3D multirater RCNN for multimodal multiclass detection and characterisation of extremely small objects. International Conference on Medical Imaging with Deep Learning,

Sudre, C. H., Van Wijnen, K., Dubost, F., Adams, H., Atkinson, D., Barkhof, F., Birhanu, M. A., Bron, E. E., Camarasa, R., Chaturvedi, N., Chen, Y., Chen, Z., Chen, S., Dou, Q., Evans, T., Ezhov, I., Gao, H., Girones Sanguesa, M., Gispert, J. D., . . . de Bruijne, M. (2024). Where is VALDO? VAscular Lesions Detection and segmentatiOn challenge at MICCAI 2021. *Medical Image Analysis*, *91*, 103029. https://doi.org/https://doi.org/10.1016/j.media.2023.103029

Tian, Y., Wang, M., Pan, Y., Meng, X., Zhao, X., Liu, L., Wang, Y., & Wang, Y. (2024). In patients who had a stroke or TIA, enlarged perivascular spaces in basal ganglia may cause future haemorrhagic strokes. *Stroke and Vascular Neurology*, *9*(1), 8. https://doi.org/10.1136/svn-2022-002157

Van Essen, D. C., Smith, S. M., Barch, D. M., Behrens, T. E. J., Yacoub, E., & Ugurbil, K. (2013). The WU-Minn Human Connectome Project: An overview. *Neuroimage*, *80*, 62-79. https://doi.org/https://doi.org/10.1016/j.neuroimage.2013.05.041

van Wijnen, K. M., Dubost, F., Yilmaz, P., Ikram, M. A., Niessen, W. J., Adams, H., Vernooij, M. W., & de Bruijne, M. (2019). Automated lesion detection by regressing intensity-based distance with a neural network. Medical Image Computing and Computer Assisted Intervention–MICCAI 2019: 22nd International Conference, Shenzhen, China, October 13–17, 2019, Proceedings, Part IV 22,

Vilor-Tejedor, N., Ciampa, I., Operto, G., Falcón, C., Suárez-Calvet, M., Crous-Bou, M., Shekari, M., Arenaza-Urquijo, E. M., Milà-Alomà, M., Grau-Rivera, O., Minguillon, C., Kollmorgen, G., Zetterberg, H., Blennow, K., Guigo, R.,

Molinuevo, J. L., & Gispert, J. D. (2021). Perivascular spaces are associated with tau pathophysiology and synaptic dysfunction in early Alzheimer's continuum. *Alzheimers Res Ther*, *13*(1), 135. https://doi.org/10.1186/s13195-021-00878-5

Vivash, L., Malpas, C. B., Meletis, C., Gollant, M., Eratne, D., Li, Q.-X., McDonald, S., O'Brien, W. T., Brodtmann, A., Darby, D., Kyndt, C., Walterfang, M., Hovens, C. M., Velakoulis, D., & O'Brien, T. J. (2022). A phase 1b open-label study of sodium selenate as a disease-modifying treatment for possible behavioral variant frontotemporal dementia. *Alzheimer's & Dementia: Translational Research & Clinical Interventions*, *8*(1), e12299. https://doi.org/https://doi.org/10.1002/trc2.12299

Wang, Y., Deng, Y., Zheng, Y., Chattopadhyay, P., & Wang, L. (2025). Vision Transformers for Image Classification: A Comparative Survey. *Technologies*, *13*(1).

Waymont, J. M. J., Valdés Hernández, M. d. C., Bernal, J., Duarte Coello, R., Brown, R., Chappell, F. M., Ballerini, L., & Wardlaw, J. M. (2024). Systematic review and meta-analysis of automated methods for quantifying enlarged perivascular spaces in the brain. *Neuroimage*, *297*, 120685. https://doi.org/https://doi.org/10.1016/j.neuroimage.2024.120685

Williamson, B. J., Khandwala, V., Wang, D., Maloney, T., Sucharew, H., Horn, P., Haverbusch, M., Alwell, K., Gangatirkar, S., Mahammedi, A., Wang, L. L., Tomsick, T., Gaskill-Shipley, M., Cornelius, R., Khatri, P., Kissela, B., & Vagal, A. (2022). Automated grading of enlarged perivascular spaces in clinical imaging data of an acute stroke cohort using an interpretable, 3D deep learning framework. *Scientific Reports*, *12*(1), 788. https://doi.org/10.1038/s41598-021-04287-4

Yang, E., Gonuguntla, V., Moon, W.-J., Moon, Y., Kim, H.-J., Park, M., & Kim, J.-H. (2021). Direct Rating Estimation of Enlarged Perivascular Spaces (EPVS) in Brain MRI Using Deep Neural Network. *Applied Sciences*, *11*(20), 9398. https://www.mdpi.com/2076-3417/11/20/9398

Yushkevich, P. A., Piven, J., Hazlett, H. C., Smith, R. G., Ho, S., Gee, J. C., & Gerig, G. (2006). User-guided 3D active contour segmentation of anatomical structures: significantly improved efficiency and reliability. *Neuroimage*, *31*(3), 1116-1128. https://doi.org/10.1016/j.neuroimage.2006.01.015

Zabihi, M., Tangwiriyasakul, C., Ingala, S., Lorenzini, L., Camarasa, R., Barkhof, F., de Bruijne, M., Cardoso, M. J., & Sudre, C. H. (2023, Oct 08). Leveraging Ellipsoid Bounding Shapes and Fast R-CNN for Enlarged Perivascular Spaces Detection and Segmentation. *Lecture Notes in Computer Science* [Machine learning in medical imaging, mlmi 2023, pt ii]. 14th International Workshop on Machine Learning in Medical Imaging (MLMI), Vancouver, CANADA.

Zhuo, J. C., Raghavan, P., Shao, M. H., Roys, S., Liang, X., Tchoquessi, R. L. N., Rhodes, C. S., Badjatia, N., Prince, J. L., & Gullapalli, R. P. (2024). Automatic Quantification of Enlarged Perivascular Space in Patients With Traumatic

Brain Injury Using Super-Resolution of T2-Weighted Images. *Journal of Neurotrauma*, *41*(3-4), 407-419. https://doi.org/10.1089/neu.2023.0082

**SUPPLEMENTARY DATA**

*S1. In-house T1w Dataset Study Participants*

The 3D T1-weighted MRI imaging ($n = 40$) for this study was sourced from four neuroimaging studies conducted in Australia and three publicly available datasets from the United States. The Australian studies included (i) participants from the Healthy Brain Ageing Program (HBA) at the Brain and Mind Centre, University of Sydney, which included older adults at risk for dementia (Kong et al., 2021); (ii) the Australian Football (AF) study on traumatic brain injury in amateur Australian rules football players (Major et al., 2021); (iii) the Fronto-Temporal Dementia (FTD) study, also from the Royal Melbourne Hospital, which involved patients participating in a phase 1b open-label trial of sodium selenate as a treatment for fronto-temporal dementia (Vivash et al., 2022); and (iv) the mild cognitive impairment study (MCIS) from Alfred Hospital, aiming to characterize the cognitive and biomarker profiles of patients with suspected mild cognitive impairment (ATRN12620001246976).

The publicly available datasets included the Alzheimer's Disease Neuroimaging Initiative (ADNI) (Jack et al., 2008), the Human Connectome Project (HCP) (Van Essen et al., 2013), and the Nathan-Kline Rockland Sample (NKI-RS) (Nooner et al., 2012). We combined the HCP and NKI-RS images ($n = 2$ each) into an 'Assorted Controls' (ASC) group due to small sample size and comparable demographics. The primary inclusion criterion for this study was the availability of 3D T1-weighted MRI images. A random selection of participants was made from each dataset, and further details regarding the individual studies, participant demographics, and MRI parameters can be found in Table S2.

*S2. Comparison between HCP-Aging T2w labels and in-house T1w labels*

To account for any differences in model performance between the T2w and T1w datasets, we sought to determine whether the underlying characteristics of the labelled PVS, such as cluster size distribution and image contrast, could account for these differences. As shown in Figure S1, the log-scaled density distribution shows a predominance of small PVS clusters across both modalities. However, a longer tail of large clusters is present in our dataset of T1w PVS segmentations. T2w images demonstrated significantly higher contrast (see Figure S2), which highlights the enhanced visibility of PVS on T2w MRI compared to T1w.

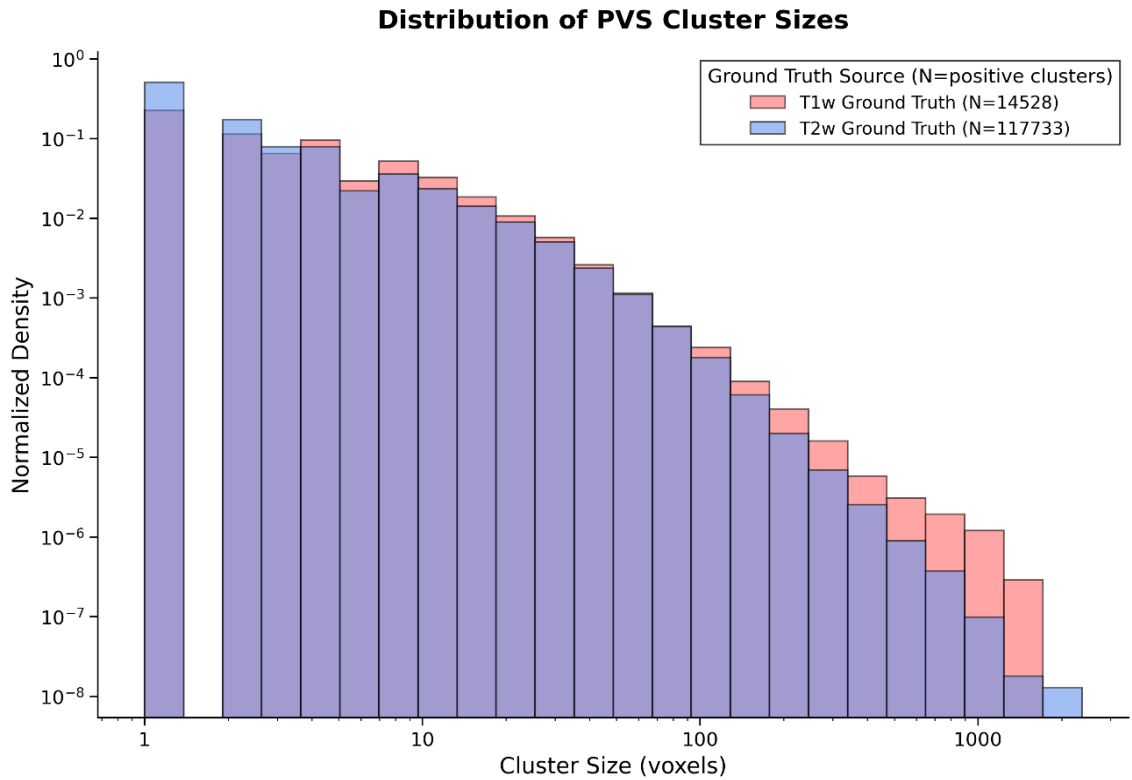

**Fig. S1:** *The log-scaled distribution of PVS cluster sizes across manually segmented T1-weighted (internal) and T2-weighted (HCP-Aging) datasets. Normalized density denotes the proportion (probability) of PVS detected at each cluster size, normalized within each modality.*

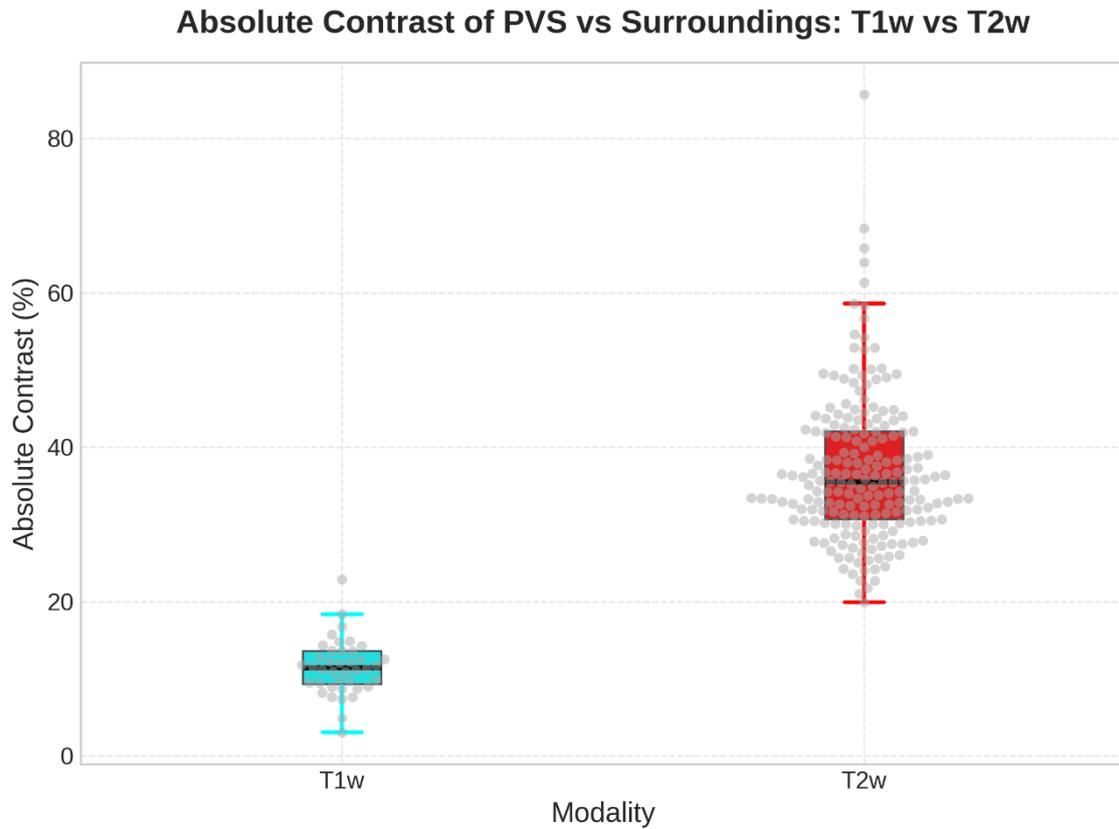

**Fig. S2:** *Absolute image contrast between PVS voxels and surrounding tissue in T1w and T2w MRI. The mean absolute intensity difference between manually segmented PVS voxels and their immediate surroundings (1-voxel dilation) across modalities was quantified.*

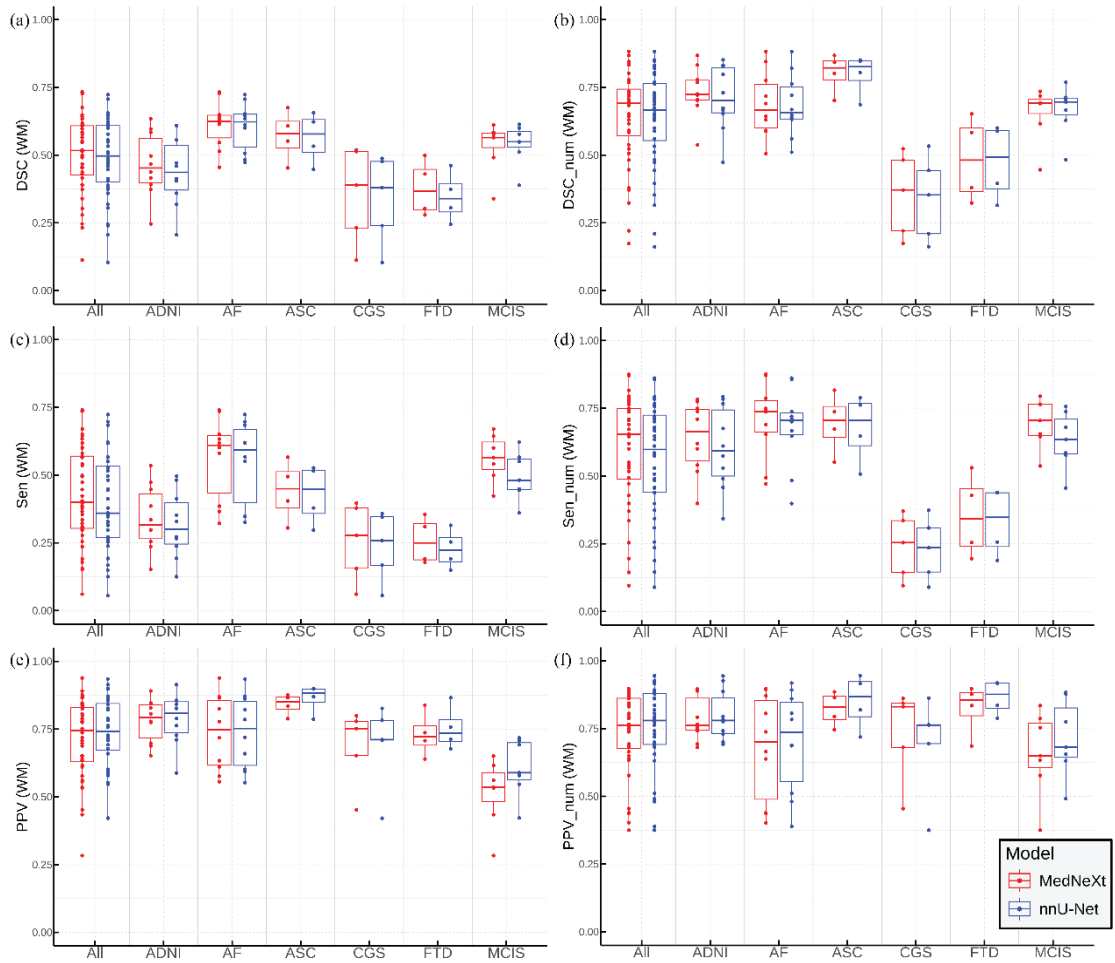

**Fig. S3**: *Performance metrics from internal 5-fold cross-validation (5FCV) in the white matter (WM) Comparison of MedNeXt-L-k5 (red) and nnU-Net (blue) trained and evaluated on data from all sites. Subplots display metrics for individual site evaluations. DSC, Dice score; Sen, sensitivity; PPV, positive predictive value; num, cluster-level metrics.*

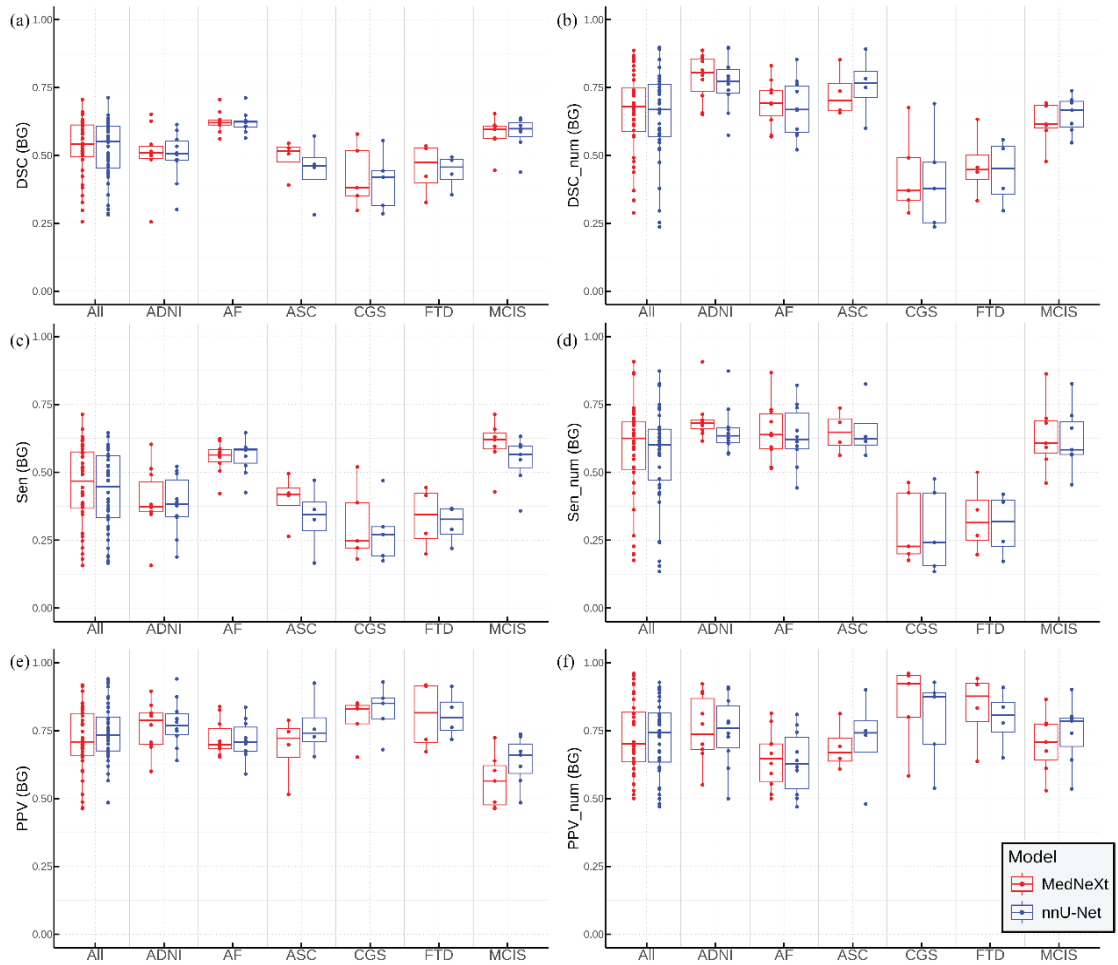

**Fig. S4**: *Performance metrics from internal 5-fold cross-validation (5FCV) in the basal ganglia (BG). Comparison of MedNeXt-L-k5 (red) and nnU-Net (blue) trained and evaluated on data from all sites. Subplots display metrics for individual site evaluations. DSC, Dice score; Sen, sensitivity; PPV, positive predictive value; num, cluster-level metrics.*

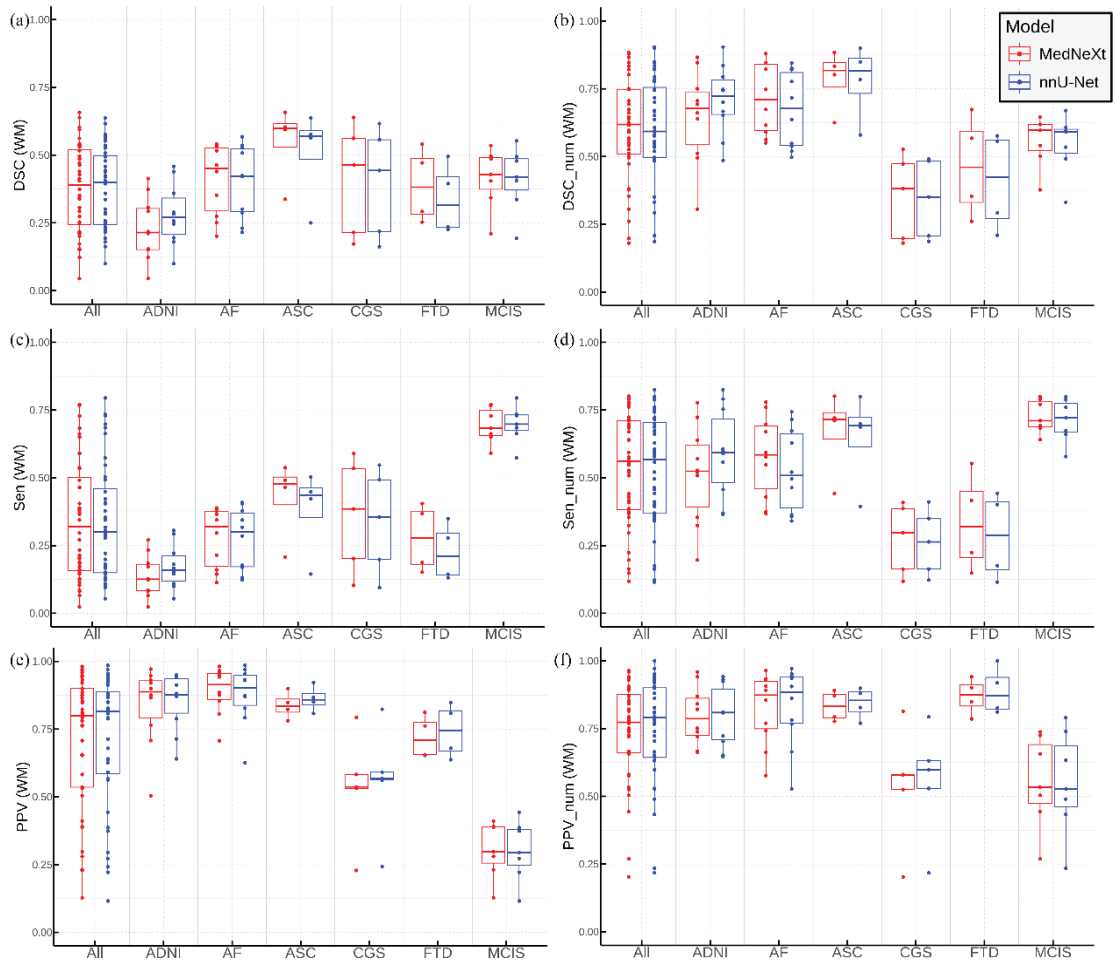

**Fig. S5:** *Performance metrics from leave-one-site-out cross validation (LOSOCV) in the white matter (WM). Subplots display metrics from individual leave-out site evaluations. DSC, Dice score; Sen, sensitivity; PPV, positive predictive value; num, cluster-level metrics.*

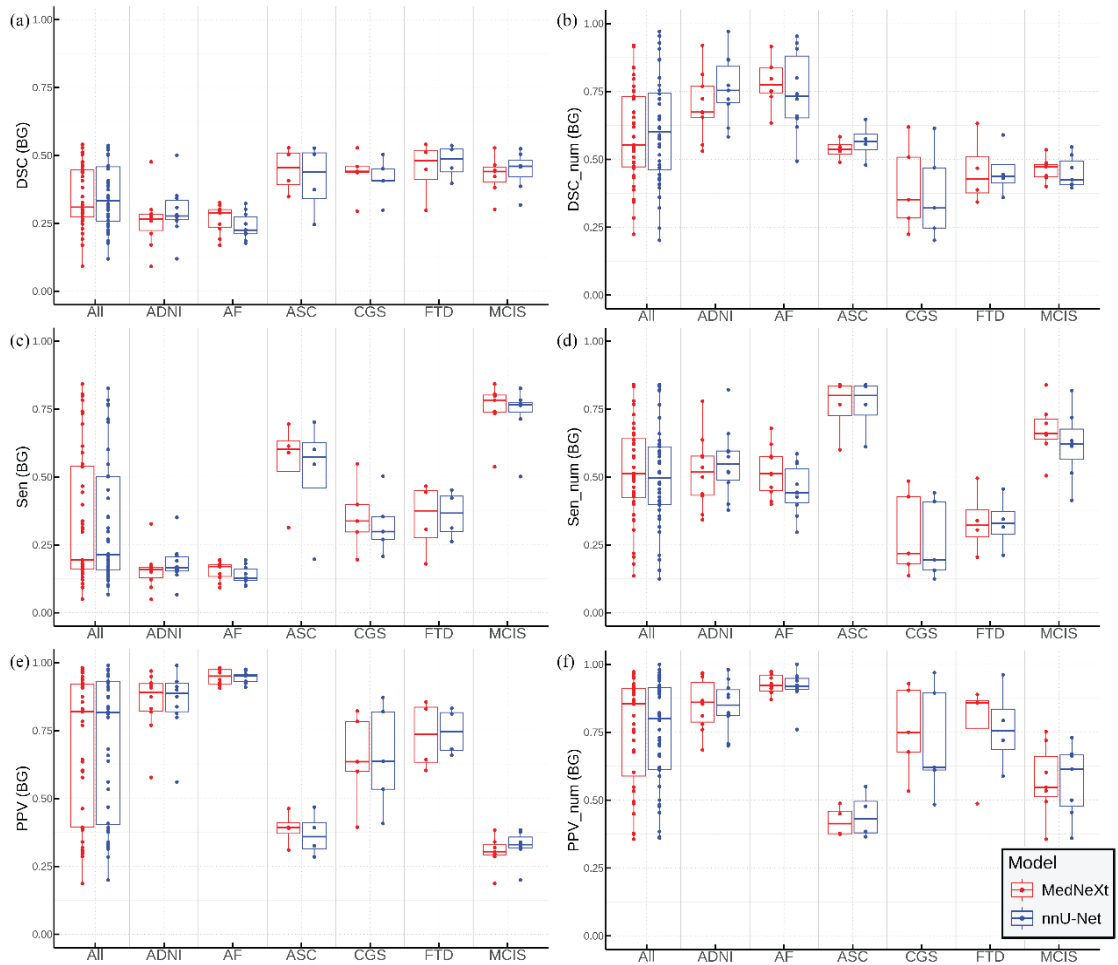

**Fig. S6**: *Performance metrics from leave-one-site-out cross validation (LOSOCV) in the basal ganglia (BG). Subplots display metrics from individual leave-out site evaluations. DSC, Dice score; Sen, sensitivity; PPV, positive predictive value; num, cluster-level metrics.*

**Table S1:** T1- and T2-weighted MRI acquisition parameters for the HCP-Aging dataset.

| Parameter | T1w | T2w |
|---|---|---|
| Scanner | Siemens 3T Prisma | Siemens 3T Prisma |
| Gradient Coil Strength | 80 mT/m | 80 mT/m |
| Receive Coil | 32-channel Siemens head coil | 32-channel Siemens head coil |
| Sequence | Multi-echo MPRAGE | Turbo spin echo (TSE) |
| Repetition Time (TR) | 2500 ms | 3200 ms |
| Inversion Time (TI) | 1000 ms | - |
| Echo Time (TE) | 1.8, 3.6, 5.4, 7.2 ms | 564 ms |
| Flip Angle | 8° | - |
| Turbo Factor | - | 314 |
| Voxel Size | 0.8 mm isotropic | 0.8 mm isotropic |
| Field of View (FOV) | 256 x 240 x 166 mm | 256 x 240 x 166 mm |
| Matrix | 320 x 300 x 208 slices | 320 x 300 x 208 slices |

***Table S2***: *An overview of the six datasets used for both manual segmentation and for training and evaluating the algorithm. The abbreviations are defined as follows: ADNI: Alzheimer's Disease Neuroimaging Initiative; AF: Australian Football study; ASC: Assorted Controls; FTD: Frontotemporal Dementia; MCIS: Mild Cognitive Impairment; AD: Alzheimer's Disease; CN: Cognitively Normal; WM: White Matter; BG: Basal Ganglia; RMH: Royal Melbourne Hospital. Numeric values are shown in the format Mean (Standard Deviation).*

| Study | ADNI | AF | ASC | HBA | FTD | MCIS |
|---|---|---|---|---|---|---|
| Number of participants | 10 | 10 | 4 | 5 | 4 | 7 |
| Location | USA - various | RMH | USA - various | IMED Radiology (Camperdown) | RMH | Alfred |
| Age | 74.1 (4.1) | 26.2 (1.8) | 21 (8.4) | 64.4 (8.3) | 61.3 (10.1) | 68.6 (6.4) |
| Condition | CN (3) MCI (4) AD (3) | AF players (4) AF concussion (1) Non-contact sport (5) | Healthy controls | Subjective cognitive decline (1) MCI (4) | Behavioral variant FTD | FTD (1) Parkinson's-plus syndrome (1) MCI-AD spectrum (5) |
| Scanner | Signa Excite (5) Symphony (3) Genesis Signa (1) Trio (1) | Siemens Magnetom Prisma | Siemens (Custom) Siemens TrioTim | General Electric Discovery MR750 | Siemens Magnetom Prisma | Siemens Magnetom Prisma |
| Field strength | 1.5T | 3T | 3T | 3T | 3T | 3T |
| Sequence | MPRAGE | MPRAGE | MPRAGE | BRAVO SPGR | MPRAGE | MPRAGE |
| Voxel size | 0.9x0.9x1.2 (5) 1.0x1.0x1.2 (1) 1.1x1.1x1.2 (1) 1.3x1.3x1.2 (3) | 0.8x0.8x0.8 | 1.0x1.0x1.0 (2) 1.0x0.98x0.98 (2) | 1.0x1.0x1.0 | 1.0x1.0x1.0 | 0.8x0.8x0.8 |
| Average number PVS (WM) | 166 (92) | 266 (135) | 218 (67) | 303 (169) | 292 (124) | 484 (311) |
| Average volume PVS (WM) | 3580 (1491) | 3162 (2023) | 2962 (1161) | 1943 (1394) | 1703 (538) | 2130 (1453) |
| Average number PVS (BG) | 32 (9) | 33 (9) | 17 (8) | 66 (27) | 50 (10) | 98 (41) |
| Average volume PVS (BG) | 1347 (617) | 957 (235) | 294 (154) | 386 (327) | 476 (137) | 385 (176) |

***Table S3:*** *Comparison of MedNeXt-L-k5 and nnU-Net performance on the HCP-Aging dataset.*

| Region | Modality | Model | Dice Score | | Sensitivity | | Precision | | Correlation | |
|---|---|---|---|---|---|---|---|---|---|---|
| | | | Voxel | Number | Voxel | Number | Voxel | Number | Voxel | Number |
| White Matter | T2w | MedNeXt-L-k5 | 0.88(0.06) | 0.88(0.09) | 0.84(0.08) | 0.86(0.13) | 0.92(0.06) | 0.90(0.07) | 0.99 | 0.88 |
| | | nnU-Net | 0.89(0.06) | 0.89(0.09) | 0.85(0.08) | 0.86(0.13) | 0.93(0.05) | 0.91(0.07) | 0.99 | 0.88 |
| | T1w | MedNeXt-L-k5 | 0.58(0.09) | 0.64(0.11) | 0.57(0.10) | 0.57(0.12) | 0.60(0.10) | 0.76(0.10) | 0.98 | 0.86 |

***Table S4:*** *Comparison of MedNeXt-L-k5 performance on internal 5FCV with and without preprocessing.*

| Region | Images used for Training & Evaluation | Dice Score | | Sensitivity | | Precision | | Correlation | |
|---|---|---|---|---|---|---|---|---|---|
| | | Voxel | Number | Voxel | Number | Voxel | Number | Voxel | Number |
| White Matter | Raw | **0.51(0.14)** | **0.64(0.18)** | **0.44(0.18)** | **0.60(0.21)** | **0.72(0.16)** | 0.73(0.15) | **0.80** | **0.76** |
| | Preprocessed | 0.49(0.14) | 0.64(0.18) | 0.40(0.16) | 0.58(0.21) | **0.72(0.16)** | **0.74(0.17)** | **0.80** | **0.76** |
| Basal Ganglia | Raw | **0.53(0.11)** | **0.65(0.16)** | **0.46(0.14)** | **0.58(0.18)** | **0.72(0.12)** | **0.73(0.13)** | **0.88** | **0.67** |
| | Preprocessed | 0.51(0.11) | 0.62(0.16) | 0.43(0.14) | 0.56(0.19) | 0.71(0.15) | 0.72(0.16) | 0.82 | 0.66 |

**Table S5**: Dice scores (voxel-level), for each algorithm on each dataset. Average (standard deviation) is reported as the mean across all subjects in validation set(s). LO: leave out, 5FCV: 5-fold cross validation.

| Region | Algorithm | Training Dataset | Internal Validation (5FCV) | External Validation | | | | | | |
|---|---|---|---|---|---|---|---|---|---|---|
| | | | | ADNI | AF | ASC | CGS | FTD | MCIS | Average |
| White Matter | SHIVA | - | - | 0.22 (0.08) | 0.05 (0.03) | 0.36 (0.08) | 0.31 (0.10) | 0.20 (0.10) | 0.08 (0.02) | 0.18 (0.13) |
| | WPSS | - | - | 0.35 (0.09) | 0.25 (0.08) | 0.35 (0.05) | 0.29 (0.12) | 0.26 (0.09) | 0.29 (0.11) | 0.30 (0.10) |
| | mc-PVS Net | - | - | 0.34 (0.13) | 0.22 (0.10) | 0.44 (0.13) | 0.38 (0.20) | 0.44 (0.12) | 0.33 (0.10) | 0.33 (0.14) |
| | nnU-Net | All Sites | 0.49 (0.15) | - | - | - | - | - | - | 0.38 (0.15) |
| | | LO ADNI | 0.50 (0.15) | 0.28 (0.11) | - | - | - | - | - | |

| Region | Method | Subset | | | | | | | | |
|---|---|---|---|---|---|---|---|---|---|---|
| | | LO AF | 0.45 (0.14) | - | 0.40 (0.13) | - | - | - | - | |
| | | LO ASC | 0.48 (0.15) | - | - | 0.51 (0.17) | - | - | - | |
| | | LO CGS | 0.52 (0.12) | - | - | - | 0.40 (0.20) | - | - | |
| | | LO FTD | 0.51 (0.15) | - | - | - | - | 0.34 (0.13) | - | |
| | | LO MCIS | 0.49 (0.16) | - | - | - | - | - | 0.41 (0.12) | |
| | MedNeXt-L-k5 | All Sites | 0.51 (0.14) | - | - | - | - | - | - | |
| | | LO ADNI | 0.52 (0.14) | 0.23 (0.12) | - | - | - | - | - | |
| | | LO AF | 0.46 (0.14) | - | 0.41 (0.13) | - | - | - | - | |
| | | LO ASC | 0.50 (0.16) | - | - | 0.55 (0.14) | - | - | - | 0.38 (0.16) |
| | | LO CGS | 0.53 (0.12) | - | - | - | 0.41 (0.21) | - | - | |
| | | LO FTD | 0.52 (0.15) | - | - | - | - | 0.39 (0.14) | - | |
| | | LO MCIS | 0.50 (0.16) | - | - | - | - | - | 0.41 (0.11) | |
| Basal Ganglia | SHIVA | - | - | 0.04 (0.03) | 0.09 (0.05) | 0.04 (0.05) | 0.35 (0.04) | 0.11 (0.10) | 0.08 (0.10) | 0.10 (0.11) |
| | WPSS | - | - | 0.16 (0.09) | 0.17 (0.04) | 0.13 (0.06) | 0.26 (0.08) | 0.16 (0.10) | 0.29 (0.05) | 0.20 (0.09) |
| | mc-PVS Net | - | - | 0.27 (0.11) | 0.15 (0.04) | 0.23 (0.20) | 0.34 (0.16) | 0.41 (0.07) | 0.40 (0.04) | 0.28 (0.13) |
| | nnU-Net | All Sites | 0.52 (0.11) | - | - | - | - | - | - | |
| | | LO ADNI | 0.52 (0.12) | 0.29 (0.09) | - | - | - | - | - | |
| | | LO AF | 0.49 (0.10) | - | 0.24 (0.05) | - | - | - | - | |
| | | LO ASC | 0.53 (0.11) | - | - | 0.41 (0.13) | - | - | - | 0.35 (0.12) |
| | | LO CGS | 0.54 (0.10) | - | - | - | 0.41 (0.08) | - | - | |
| | | LO FTD | 0.54 (0.11) | - | - | - | - | 0.48 (0.06) | - | |
| | | LO MCIS | 0.52 (0.11) | - | - | - | - | - | 0.44 (0.07) | |
| | MedNeXt-L-k5 | All Sites | 0.53 (0.11) | - | - | - | - | - | - | |
| | | LO ADNI | 0.53 (0.11) | 0.26 (0.10) | - | - | - | - | - | |
| | | LO AF | 0.49 (0.10) | - | 0.27 (0.05) | - | - | - | - | |
| | | LO ASC | 0.54 (0.11) | - | - | 0.45 (0.08) | - | - | - | 0.35 (0.12) |
| | | LO CGS | 0.55 (0.10) | - | - | - | 0.43 (0.09) | - | - | |
| | | LO FTD | 0.54 (0.11) | - | - | - | - | 0.45 (0.11) | - | |
| | | LO MCIS | 0.52 (0.11) | - | - | - | - | - | 0.43 (0.07) | |

**Table S6**: Dice scores (cluster-level), for each algorithm on each dataset. Average (standard deviation) is reported as the mean across all subjects in validation set(s). LO: leave out, 5FCV: 5-fold cross validation.

| Region | Algorithm | Training Dataset | Internal Validation (5FCV) | External Validation | | | | | | |
|--------|-----------|------------------|----------------------------|------|------|------|------|------|------|---------|
| | | | | ADNI | AF | ASC | CGS | FTD | MCIS | Average |
| White Matter | SHIVA | - | - | 0.49 (0.10) | 0.13 (0.07) | 0.56 (0.09) | 0.30 (0.10) | 0.25 (0.14) | 0.13 (0.04) | 0.29 (0.19) |
| | WPSS | - | - | 0.51 (0.16) | 0.27 (0.11) | 0.64 (0.10) | 0.26 (0.05) | 0.41 (0.10) | 0.27 (0.15) | 0.38 (0.18) |
| | mc-PVS Net | - | - | 0.64(0.12) | 0.45(0.15) | 0.63(0.15) | 0.31(0.14) | 0.47(0.16) | 0.42(0.14) | 0.50(0.17) |
| | nnU-Net | All Sites | 0.64 (0.18) | - | - | - | - | - | - | |
| | | LO ADNI | 0.61 (0.19) | 0.71 (0.13) | - | - | - | - | - | |
| | | LO AF | 0.61 (0.20) | - | 0.67 (0.14) | - | - | - | - | |
| | | LO ASC | 0.62 (0.18) | - | - | 0.78 (0.14) | - | - | - | 0.60 (0.19) |
| | | LO CGS | 0.69 (0.13) | - | - | - | 0.34 (0.15) | - | - | |
| | | LO FTD | 0.65 (0.18) | - | - | - | - | 0.41 (0.19) | - | |
| | | LO MCIS | 0.63 (0.19) | - | - | - | - | - | 0.55 (0.11) | |
| | MedNeXt-L-k5 | All Sites | 0.64 (0.18) | - | - | - | - | - | - | |
| | | LO ADNI | 0.62 (0.18) | 0.65 (0.17) | - | - | - | - | - | |
| | | LO AF | 0.64 (0.19) | - | 0.71 (0.13) | - | - | - | - | |
| | | LO ASC | 0.62 (0.18) | - | - | 0.79 (0.11) | - | - | - | 0.61 (0.19) |
| | | LO CGS | 0.69 (0.14) | - | - | - | 0.36 (0.21) | - | - | |
| | | LO FTD | 0.66 (0.18) | - | - | - | - | 0.46 (0.19) | - | |
| | | LO MCIS | 0.63 (0.19) | - | - | - | - | - | 0.55(0.17) | |
| Basal Ganglia | SHIVA | - | - | 0.16 (0.11) | 0.20 (0.07) | 0.12 (0.11) | 0.19 (0.08) | 0.12 (0.07) | 0.09 (0.07) | 0.15 (0.09) |
| | WPSS | - | - | 0.53 (0.23) | 0.38 (0.05) | 0.29 (0.16) | 0.30 (0.06) | 0.32 (0.14) | 0.28 (0.08) | 0.37 (0.17) |
| | mc-PVS Net | - | - | 0.73 (0.19) | 0.56 (0.09) | 0.39 (0.27) | 0.39 (0.18) | 0.50 (0.09) | 0.46 (0.04) | 0.54 (0.19) |
| | nnU-Net | All Sites | 0.64 (0.17) | - | - | - | - | - | - | |
| | | LO ADNI | 0.60 (0.14) | 0.76 (0.12) | - | - | - | - | - | 0.60 (0.20) |
| | | LO AF | 0.63 (0.18) | - | 0.75 (0.15) | - | - | - | - | |
| | | LO ASC | 0.65 (0.18) | - | - | 0.56 (0.07) | - | - | - | |

| | Training Dataset | Internal Validation | ADNI | AF | ASC | CGS | FTD | MCIS | Average |
|---|---|---|---|---|---|---|---|---|---|
| | LO CGS | 0.68 (0.13) | - | - | - | 0.37 (0.17) | - | - | |
| | LO FTD | 0.67 (0.14) | - | - | - | - | 0.46 (0.10) | - | |
| | LO MCIS | 0.64 (0.18) | - | - | - | - | - | 0.45 (0.06) | |
| MedNeXt-L-k5 | All Sites | 0.65 (0.16) | - | - | - | - | - | - | |
| | LO ADNI | 0.59 (0.14) | 0.73 (0.15) | - | - | - | - | - | |
| | LO AF | 0.63 (0.17) | - | 0.83 (0.13) | - | - | - | - | |
| | LO ASC | 0.63 (0.17) | - | - | 0.54 (0.04) | - | - | - | 0.62 (0.21) |
| | LO CGS | 0.69 (0.14) | - | - | - | 0.36 (0.13) | - | - | |
| | LO FTD | 0.66 (0.14) | - | - | - | - | 0.46 (0.13) | - | |
| | LO MCIS | 0.64 (0.16) | - | - | - | - | - | 0.46 (0.04) | |

**Table S7**: Sensitivity (Recall; voxel-level), for each algorithm on each dataset. Average (standard deviation) is reported as the mean across all subjects in validation set(s). LO: leave out, 5FCV: 5-fold cross validation.

| Region | Algorithm | Training Dataset | Internal Validation (5FCV) | External Validation | | | | | | |
|---|---|---|---|---|---|---|---|---|---|---|
| | | | | ADNI | AF | ASC | CGS | FTD | MCIS | Average |
| White Matter | SHIVA | - | - | 0.14 (0.07) | 0.03 (0.02) | 0.24 (0.07) | 0.26 (0.13) | 0.12 (0.07) | 0.04 (0.01) | 0.12 (0.10) |
| | WPSS | - | - | 0.28 (0.08) | 0.22 (0.06) | 0.25 (0.04) | 0.28 (0.16) | 0.18 (0.08) | 0.38 (0.06) | 0.27 (0.10) |
| | mc-PVS Net | - | - | 0.22 (0.10) | 0.13 (0.06) | 0.29 (0.10) | 0.34 (0.19) | 0.35 (0.12) | 0.26 (0.10) | 0.24 (0.13) |
| | nnU-Net | All Sites | 0.40 (0.17) | - | - | - | - | - | - | |
| | | LO ADNI | 0.43 (0.18) | 0.17 (0.08) | - | - | - | - | - | |
| | | LO AF | 0.34 (0.14) | - | 0.27 (0.11) | - | - | - | - | |
| | | LO ASC | 0.40 (0.18) | - | - | 0.38 (0.16) | - | - | - | 0.34 (0.21) |
| | | LO CGS | 0.42 (0.15) | - | - | - | 0.34 (0.19) | - | - | |
| | | LO FTD | 0.43 (0.17) | - | - | - | - | 0.23 (0.11) | - | |
| | | LO MCIS | 0.39 (0.18) | - | - | - | - | - | 0.70 (0.07) | |
| | MedNeXt-L-k5 | All Sites | 0.44 (0.18) | - | - | - | - | - | - | |
| | | LO ADNI | 0.46 (0.17) | 0.14 (0.08) | - | - | - | - | - | 0.34 (0.22) |
| | | LO AF | 0.36 (0.15) | - | 0.28 (0.11) | - | - | - | - | |

| Region | Algorithm | Training Dataset | Internal Validation (5FCV) | ADNI | AF | ASC | CGS | FTD | MCIS | Average |
|---|---|---|---|---|---|---|---|---|---|---|
| | | LO ASC | 0.43 (0.19) | - | - | 0.43 (0.15) | - | - | - | |
| | | LO CGS | 0.47 (0.16) | - | - | - | 0.36 (0.21) | - | - | |
| | | LO FTD | 0.45 (0.18) | - | - | - | - | 0.28 (0.13) | - | |
| | | LO MCIS | 0.41 (0.18) | - | - | - | - | - | 0.69 (0.07) | |
| Basal Ganglia | SHIVA | - | - | 0.02 (0.02) | 0.05 (0.03) | 0.03 (0.04) | 0.26 (0.04) | 0.07 (0.06) | 0.05 (0.08) | 0.07 (0.09) |
| | WPSS | - | - | 0.10 (0.06) | 0.11 (0.03) | 0.10 (0.04) | 0.21 (0.07) | 0.10 (0.07) | 0.31 (0.08) | 0.15 (0.10) |
| | mc-PVS Net | - | - | 0.17 (0.08) | 0.08 (0.02) | 0.16 (0.15) | 0.26 (0.14) | 0.30 (0.08) | 0.33 (0.05) | 0.20 (0.12) |
| | nnU-Net | All Sites | 0.43 (0.14) | - | - | - | - | - | - | |
| | | LO ADNI | 0.44 (0.15) | 0.18 (0.07) | - | | | | | |
| | | LO AF | 0.38 (0.13) | - | 0.14 (0.03) | - | | | | |
| | | LO ASC | 0.44 (0.14) | - | - | 0.51 (0.22) | - | | | 0.34 (0.24) |
| | | LO CGS | 0.45 (0.13) | - | - | - | 0.33 (0.11) | - | | |
| | | LO FTD | 0.45 (0.14) | - | - | - | - | 0.36 (0.09) | - | |
| | | LO MCIS | 0.42 (0.14) | - | - | - | - | - | 0.73 (0.11) | |
| | MedNeXt-L-k5 | All Sites | 0.46 (0.14) | - | - | - | - | - | - | |
| | | LO ADNI | 0.47 (0.15) | 0.16 (0.07) | - | | | | | |
| | | LO AF | 0.40 (0.14) | - | 0.16 (0.03) | - | | | | |
| | | LO ASC | 0.47 (0.15) | - | - | 0.55 (0.17) | - | | | 0.34 (0.25) |
| | | LO CGS | 0.48 (0.14) | - | - | - | 0.36 (0.13) | - | | |
| | | LO FTD | 0.47 (0.15) | - | - | - | - | 0.35 (0.13) | - | |
| | | LO MCIS | 0.43 (0.14) | - | - | - | - | - | 0.75 (0.10) | |

**Table S8**: Sensitivity (Recall; cluster-level), for each algorithm on each dataset. Average (standard deviation) is reported as the mean across all subjects in validation set(s). LO: leave out, 5FCV: 5-fold cross validation.

| Region | Algorithm | Training Dataset | Internal Validation (5FCV) | External Validation | | | | | | |
|---|---|---|---|---|---|---|---|---|---|---|
| | | | | ADNI | AF | ASC | CGS | FTD | MCIS | Average |
| White Matter | SHIVA | - | - | 0.38 (0.13) | 0.07 (0.04) | 0.38 (0.07) | 0.22 (0.10) | 0.16 (0.10) | 0.06 (0.02) | 0.21 (0.16) |
| | WPSS | - | - | 0.80 (0.09) | 0.77 (0.09) | 0.73 (0.07) | 0.35 (0.16) | 0.37 (0.18) | 0.77 (0.09) | 0.68 (0.21) |

| Region | Model | Site | | | | | | | | |
|---|---|---|---|---|---|---|---|---|---|---|
| | mc-PVS Net | - | - | 0.51 (0.12) | 0.29 (0.11) | 0.44 (0.14) | 0.23 (0.11) | 0.36 (0.17) | 0.30 (0.12) | 0.36 (0.16) |
| | nnU-Net | All Sites | 0.57 (0.21) | - | - | - | - | - | - | |
| | | LO ADNI | 0.56 (0.23) | 0.59 (0.16) | - | - | - | - | - | |
| | | LO AF | 0.52 (0.21) | - | 0.53 (0.15) | - | - | - | - | |
| | | LO ASC | 0.56 (0.22) | - | - | 0.65 (0.17) | - | - | - | 0.53 (0.21) |
| | | LO CGS | 0.63 (0.16) | - | - | - | 0.26 (0.12) | - | - | |
| | | LO FTD | 0.60 (0.20) | - | - | - | - | 0.28 (0.16) | - | |
| | | LO MCIS | 0.55 (0.22) | - | - | - | - | - | 0.71 (0.08) | |
| | MedNeXt-L-k5 | All Sites | 0.60 (0.21) | - | - | - | - | - | - | |
| | | LO ADNI | 0.59 (0.22) | 0.51 (0.18) | - | - | - | - | - | |
| | | LO AF | 0.55 (0.21) | - | 0.58 (0.15) | - | - | - | - | |
| | | LO ASC | 0.58 (0.22) | - | - | 0.67 (0.16) | - | - | - | 0.54 (0.20) |
| | | LO CGS | 0.66 (0.16) | - | - | - | 0.27 (0.13) | - | - | |
| | | LO FTD | 0.63 (0.21) | - | - | - | - | 0.34 (0.18) | - | |
| | | LO MCIS | 0.57 (0.22) | - | - | - | - | - | 0.73 (0.06) | |
| Basal Ganglia | SHIVA | - | - | 0.09 (0.06) | 0.11 (0.04) | 0.08 (0.08) | 0.11 (0.04) | 0.06 (0.04) | 0.05 (0.04) | 0.09 (0.05) |
| | WPSS | - | - | 0.53 (0.19) | 0.76 (0.06) | 0.58 (0.11) | 0.34 (0.05) | 0.33 (0.17) | 0.60 (0.11) | 0.56 (0.19) |
| | mc-PVS Net | - | - | 0.51 (0.12) | 0.29 (0.11) | 0.44 (0.14) | 0.23 (0.11) | 0.36 (0.17) | 0.30 (0.12) | 0.37 (0.15) |
| | nnU-Net | All Sites | 0.56 (0.18) | - | - | - | - | - | - | |
| | | LO ADNI | 0.52 (0.19) | 0.55 (0.13) | - | - | - | - | - | |
| | | LO AF | 0.53 (0.19) | - | 0.45 (0.09) | - | - | - | - | |
| | | LO ASC | 0.56 (0.19) | - | - | 0.76 (0.11) | - | - | - | 0.50 (0.18) |
| | | LO CGS | 0.60 (0.14) | - | - | - | 0.27 (0.15) | - | - | |
| | | LO FTD | 0.60 (0.17) | - | - | - | - | 0.33 (0.10) | - | |
| | | LO MCIS | 0.55 (0.20) | - | - | - | - | - | 0.62 (0.13) | |
| | MedNeXt-L-k5 | All Sites | 0.58 (0.18) | - | - | - | - | - | - | |
| | | LO ADNI | 0.54 (0.19) | 0.52 (0.13) | - | - | - | - | - | |
| | | LO AF | 0.55 (0.20) | - | 0.52 (0.09) | - | - | - | - | |
| | | LO ASC | 0.56 (0.19) | - | - | 0.76 (0.11) | - | - | - | 0.52 (0.18) |
| | | LO CGS | 0.63 (0.16) | - | - | - | 0.29 (0.16) | - | - | |
| | | LO FTD | 0.60 (0.16) | - | - | - | - | 0.34 (0.12) | - | |

| | | | | | | | |
|---|---|---|---|---|---|---|---|
| LO MCIS | 0.56 (0.19) | - | - | - | - | - | 0.67 (0.10) |

**Table S9**: Specificity (voxel-level), for each algorithm on each dataset. Average (standard deviation) is reported as the mean across all subjects in validation set(s). LO: leave out, 5FCV: 5-fold cross validation.

| Region | Algorithm | Training Dataset | Internal Validation (5FCV) | External Validation | | | | | | |
|---|---|---|---|---|---|---|---|---|---|---|
| | | | | ADNI | AF | ASC | CGS | FTD | MCIS | Average |
| White Matter | SHIVA | - | - | 0.71 (0.11) | 0.87 (0.04) | 0.77 (0.05) | 0.46 (0.07) | 0.68 (0.11) | 0.43 (0.14) | 0.67 (0.19) |
| | WPSS | - | - | 0.50 (0.15) | 0.30 (0.14) | 0.60 (0.08) | 0.36 (0.09) | 0.55 (0.08) | 0.25 (0.13) | 0.40 (0.17) |
| | mc-PVS Net | - | - | 0.79 (0.15) | 0.88 (0.05) | 0.94 (0.03) | 0.47 (0.21) | 0.63 (0.08) | 0.50 (0.05) | 0.72 (0.20) |
| | nnU-Net | All Sites | 0.74 (0.14) | - | - | - | - | - | - | 0.71 (0.25) |
| | | LO ADNI | 0.72 (0.17) | 0.85 (0.11) | - | - | - | - | - | |
| | | LO AF | 0.75 (0.14) | - | 0.88 (0.11) | - | - | - | - | |
| | | LO ASC | 0.73 (0.14) | - | - | 0.86 (0.05) | - | - | - | |
| | | LO CGS | 0.75 (0.14) | - | - | - | 0.56 (0.21) | - | - | |
| | | LO FTD | 0.74 (0.16) | - | - | - | - | 0.74 (0.10) | - | |
| | | LO MCIS | 0.77 (0.13) | - | - | - | - | - | 0.30 (0.11) | |
| | MedNeXt-L-k5 | All Sites | 0.72 (0.16) | - | - | - | - | - | - | 0.71 (0.25) |
| | | LO ADNI | 0.69 (0.17) | 0.84 (0.14) | - | - | - | - | - | |
| | | LO AF | 0.74 (0.13) | - | 0.89 (0.09) | - | - | - | - | |
| | | LO ASC | 0.70 (0.15) | - | - | 0.84 (0.05) | - | - | - | |
| | | LO CGS | 0.71 (0.15) | - | - | - | 0.53 (0.20) | - | - | |
| | | LO FTD | 0.72 (0.15) | - | - | - | - | 0.72 (0.08) | - | |
| | | LO MCIS | 0.75 (0.12) | - | - | - | - | - | 0.30 (0.10) | |
| Basal Ganglia | SHIVA | - | - | 0.64 (0.32) | 0.73 (0.15) | 0.05 (0.07) | 0.57 (0.08) | 0.80 (0.21) | 0.62 (0.21) | 0.61 (0.28) |
| | WPSS | - | - | 0.55 (0.22) | 0.34 (0.07) | 0.22 (0.15) | 0.40 (0.14) | 0.39 (0.16) | 0.30 (0.07) | 0.39 (0.17) |
| | mc-PVS Net | - | - | 0.89 (0.05) | 0.85 (0.04) | 0.73 (0.16) | 0.59 (0.18) | 0.67 (0.02) | 0.53 (0.09) | 0.74 (0.17) |
| | nnU-Net | All Sites | 0.74 (0.10) | - | - | - | - | - | - | 0.70 (0.26) |

| Algorithm | | Internal Validation | ADNI | AF | ASC | CGS | FTD | MCIS | Average |
|---|---|---|---|---|---|---|---|---|---|
| | LO ADNI | 0.72 (0.10) | 0.86 (0.12) | - | - | - | - | - | |
| | LO AF | 0.76 (0.10) | - | 0.95 (0.02) | - | - | - | - | |
| | LO ASC | 0.75 (0.10) | - | - | 0.37 (0.08) | - | - | - | |
| | LO CGS | 0.74 (0.09) | - | - | - | 0.65 (0.19) | - | - | |
| | LO FTD | 0.74 (0.10) | - | - | - | - | 0.75 (0.09) | - | |
| | LO MCIS | 0.77 (0.09) | - | - | - | - | - | 0.32 (0.06) | |
| MedNeXt-L-k5 | All Sites | 0.72 (0.12) | - | - | - | - | - | - | |
| | LO ADNI | 0.69 (0.14) | 0.85 (0.12) | - | - | - | - | - | |
| | LO AF | 0.74 (0.11) | - | 0.95 (0.03) | - | - | - | - | |
| | LO ASC | 0.72 (0.12) | - | - | 0.39 (0.06) | - | - | - | 0.70 (0.26) |
| | LO CGS | 0.71 (0.12) | - | - | - | 0.65 (0.17) | - | - | |
| | LO FTD | 0.71 (0.12) | - | - | - | - | 0.73 (0.13) | - | |
| | LO MCIS | 0.75 (0.10) | - | - | - | - | - | 0.30 (0.06) | |

**Table S10**: Specificity (cluster-level), for each algorithm on each dataset. Average (standard deviation) is reported as the mean across all subjects in validation set(s). LO: leave out, 5FCV: 5-fold cross validation.

| Region | Algorithm | Training Dataset | Internal Validation (5FCV) | External Validation | | | | | | |
|---|---|---|---|---|---|---|---|---|---|---|
| | | | | ADNI | AF | ASC | CGS | FTD | MCIS | Average |
| White Matter | SHIVA | - | - | 0.71 (0.16) | 0.98 (0.03) | 0.91 (0.04) | 0.51 (0.06) | 0.81 (0.11) | 0.66 (0.21) | 0.77 (0.20) |
| | WPSS | - | - | 0.35 (0.15) | 0.15 (0.07) | 0.48 (0.08) | 0.24 (0.08) | 0.52 (0.09) | 0.17 (0.11) | 0.28 (0.17) |
| | mc-PVS Net | - | - | 0.80 (0.14) | 0.94 (0.09) | 0.95 (0.02) | 0.53 (0.21) | 0.77 (0.05) | 0.78 (0.12) | 0.81 (0.17) |
| | nnU-Net | All Sites | 0.76 (0.15) | - | - | - | - | - | - | |
| | | LO ADNI | 0.74 (0.18) | 0.79 (0.11) | - | - | - | - | - | |
| | | LO AF | 0.77 (0.15) | - | 0.83 (0.15) | - | - | - | - | |
| | | LO ASC | 0.75 (0.16) | - | - | 0.84 (0.06) | - | - | - | 0.75 (0.19) |
| | | LO CGS | 0.77 (0.15) | - | - | - | 0.55 (0.21) | - | - | |
| | | LO FTD | 0.75 (0.17) | - | - | - | - | 0.89 (0.09) | - | |
| | | LO MCIS | 0.77 (0.16) | - | - | - | - | - | 0.55 (0.19) | |
| | MedNeXt-L-k5 | All Sites | 0.73 (0.15) | - | - | - | - | - | - | |
| | | LO ADNI | 0.71 (0.17) | 0.80 (0.11) | - | - | - | - | - | |
| | | LO AF | 0.76 (0.12) | - | 0.82 (0.13) | - | - | - | - | |
| | | LO ASC | 0.73 (0.16) | - | - | 0.83 (0.06) | - | - | - | 0.74 (0.18) |
| | | LO CGS | 0.73 (0.16) | - | - | - | 0.54 (0.22) | - | - | |
| | | LO FTD | 0.72 (0.15) | - | - | - | - | 0.87 (0.07) | - | |
| | | LO MCIS | 0.57 (0.22) | - | - | - | - | - | 0.55 (0.17) | |
| Basal Ganglia | SHIVA | - | - | 0.64 (0.36) | 0.98 (0.06) | 0.24 (0.20) | 0.83 (0.15) | 0.95 (0.10) | 0.90 (0.10) | 0.79 (0.30) |
| | WPSS | - | - | 0.42 (0.17) | 0.21 (0.03) | 0.19 (0.12) | 0.29 (0.16) | 0.31 (0.10) | 0.18 (0.06) | 0.28 (0.15) |
| | mc-PVS Net | - | - | 0.87 (0.06) | 0.97 (0.04) | 0.68 (0.29) | 0.77 (0.19) | 0.76 (0.13) | 0.82 (0.07) | 0.85 (0.14) |
| | nnU-Net | All Sites | 0.72 (0.14) | - | - | - | - | - | - | |
| | | LO ADNI | 0.72 (0.12) | 0.85 (0.09) | - | - | - | - | - | |
| | | LO AF | 0.76 (0.13) | - | 0.92 (0.06) | - | - | - | - | |
| | | LO ASC | 0.74 (0.13) | - | - | 0.44 (0.09) | - | - | - | 0.75 (0.19) |
| | | LO CGS | 0.73 (0.13) | - | - | - | 0.72 (0.21) | - | - | |
| | | LO FTD | 0.73 (0.12) | - | - | - | - | 0.77 (0.16) | - | |
| | | LO MCIS | 0.73 (0.13) | - | - | - | - | - | 0.57 (0.13) | |
| | MedNeXt-L-k5 | All Sites | 0.73 (0.13) | - | - | - | - | - | - | |
| | | LO ADNI | 0.70 (0.16) | 0.85 (0.09) | - | - | - | - | - | |
| | | LO AF | 0.75 (0.13) | - | 0.93 (0.04) | - | - | - | - | |
| | | LO ASC | 0.74 (0.12) | - | - | 0.42 (0.06) | - | - | - | 0.76 (0.20) |
| | | LO CGS | 0.72 (0.13) | - | - | - | 0.76 (0.16) | - | - | |
| | | LO FTD | 0.72 (0.14) | - | - | - | - | 0.77 (0.19) | - | |
| | | LO MCIS | 0.56 (0.19) | - | - | - | - | - | 0.57 (0.14) | |

**Table S11:** Paired Wilcoxon signed-rank test comparing segmentation performance between MedNeXt-L-k5 and nnU-Net for the in-house T1-weighted dataset, evaluated on white matter (WM) and basal ganglia (BG). Values are medians across subjects for each model, the median difference (MedNeXt − nnU-Net), false discovery rate (FDR)–adjusted $p$-values ($p_{FDR}$ significance after FDR correction, and rank-biserial correlation ($r$) as an effect size measure. Positive median differences and positive $r$ indicate higher scores for MedNeXt-L-k5, whereas negative values indicate higher scores for nnU-Net.

| Brain region | Metric | n | Median (MedNeXt) | Median (nnU-Net) | Median diff | $P_{FDR}$ | Sig (FDR) | r |
|---|---|---|---|---|---|---|---|---|
| WM | DSC | 40 | 0.536 | 0.502 | 0.018 | 0.005 | Yes | 0.549 |
| WM | Sen | 40 | 0.431 | 0.397 | 0.034 | <0.001 | Yes | 0.778 |
| WM | PPV | 40 | 0.752 | 0.764 | -0.018 | 0.005 | Yes | -0.573 |
| WM | DSC_num | 40 | 0.690 | 0.662 | 0.005 | 0.413 | No | 0.161 |
| WM | Sen_num | 40 | 0.674 | 0.637 | 0.026 | <0.001 | Yes | 0.766 |
| WM | PPV_num | 40 | 0.772 | 0.787 | -0.024 | 0.005 | Yes | -0.556 |
| BG | DSC | 40 | 0.542 | 0.552 | 0.003 | 0.309 | No | 0.220 |
| BG | Sen | 40 | 0.468 | 0.447 | 0.026 | 0.017 | Yes | 0.461 |
| BG | PPV | 40 | 0.709 | 0.734 | -0.012 | 0.035 | Yes | -0.410 |
| BG | DSC_num | 40 | 0.679 | 0.669 | 0.002 | 0.413 | No | 0.151 |
| BG | Sen_num | 40 | 0.625 | 0.601 | 0.022 | 0.013 | Yes | 0.531 |
| BG | PPV_num | 40 | 0.701 | 0.743 | 0.013 | 0.413 | No | 0.168 |

**Table S12:** Paired Wilcoxon signed-rank test comparing segmentation performance between MedNeXt-L-k5 and nnU-Net for the HCP-Aging T2-weighted dataset, evaluated on white matter (WM). Values are medians across subjects for each model, the median difference (MedNeXt − nnU-Net), false discovery rate (FDR)–adjusted $p$-values ($p_{FDR}$) significance after FDR correction, and rank-biserial correlation ($r$) as an effect size measure. Positive median differences and positive $r$ indicate higher scores for MedNeXt-L-k5, whereas negative values indicate higher scores for nnU-Net.

| Brain region | Metric | n | Median (MedNeXt) | Median (nnU-Net) | Median diff | $P_{FDR}$ | Sig (FDR) | r |
|---|---|---|---|---|---|---|---|---|
| WM | DSC | 40 | 0.536 | 0.502 | 0.018 | 0.005 | Yes | 0.549 |
| WM | Sen | 40 | 0.431 | 0.397 | 0.034 | <0.001 | Yes | 0.778 |
| WM | PPV | 40 | 0.752 | 0.764 | -0.018 | 0.005 | Yes | -0.573 |
| WM | DSC_num | 40 | 0.690 | 0.662 | 0.005 | 0.413 | No | 0.161 |
| WM | Sen_num | 40 | 0.674 | 0.637 | 0.026 | <0.001 | Yes | 0.766 |
| WM | PPV_num | 40 | 0.772 | 0.787 | -0.024 | 0.005 | Yes | -0.556 |
| Brain region | Metric | n | Median (MedNeXt) | Median (nnU-Net) | Median diff | $P_{FDR}$ | Sig (FDR) | r |
| WM | DSC | 200 | 0.887 | 0.894 | -0.010 | <0.001 | Yes | -0.948 |
| WM | Sen | 200 | 0.860 | 0.872 | -0.007 | <0.001 | Yes | -0.605 |
| WM | PPV | 200 | 0.927 | 0.936 | -0.010 | <0.001 | Yes | -0.664 |
| WM | DSC_num | 200 | 0.913 | 0.923 | -0.009 | <0.001 | Yes | -0.676 |
| WM | Sen_num | 200 | 0.918 | 0.920 | -0.002 | 0.041 | Yes | -0.166 |
| WM | PPV_num | 200 | 0.911 | 0.928 | -0.017 | <0.001 | Yes | -0.773 |

**Table S13:** Paired Wilcoxon signed-rank test comparing segmentation performance between mcPVSNet and LOSOCV-trained models (MedNeXt-L-k5 and nnU-Net) for the in-house T1-weighted dataset, evaluated on white matter (WM) and basal ganglia (BG). Values are medians across subjects for each model, the median difference (Model A - Model B), false discovery rate (FDR)-adjusted p-values (pFDR) with significance after FDR correction, and rank-biserial correlation (r) as an effect size measure. Positive median differences and positive r indicate higher scores for Model A, whereas negative values indicate higher scores for Model B.

| Comparison | Brain region | Metric | n | Median (A) | Median (B) | Median diff (A-B) | P_FDR | Sig (FDR) |
|---|---|---|---|---|---|---|---|---|
| mcPVSnet vs. MedNeXt-L-k5 | WM | DSC | 40 | 0.339 | 0.389 | -0.065 | 0.041 | Yes |
| | WM | Sen | 40 | 0.215 | 0.320 | -0.069 | 0.007 | Yes |
| | WM | PPV | 40 | 0.790 | 0.800 | -0.008 | 0.873 | No |
| | WM | DSC_num | 40 | 0.503 | 0.618 | -0.108 | <0.001 | Yes |
| | WM | Sen_num | 40 | 0.353 | 0.561 | -0.208 | <0.001 | Yes |
| | WM | PPV_num | 40 | 0.844 | 0.773 | 0.066 | 0.004 | Yes |
| | BG | DSC | 40 | 0.277 | 0.309 | -0.062 | <0.001 | Yes |
| | BG | Sen | 40 | 0.166 | 0.195 | -0.078 | <0.001 | Yes |
| | BG | PPV | 40 | 0.811 | 0.823 | 0.006 | 0.368 | No |
| | BG | DSC_num | 40 | 0.507 | 0.601 | -0.046 | 0.015 | Yes |
| | BG | Sen_num | 40 | 0.364 | 0.512 | -0.129 | <0.001 | Yes |
| | BG | PPV_num | 40 | 0.879 | 0.857 | 0.036 | 0.003 | Yes |
| mcPVSnet vs nnU-Net | WM | DSC | 40 | 0.339 | 0.400 | -0.044 | 0.040 | Yes |
| | WM | Sen | 40 | 0.215 | 0.299 | -0.060 | 0.016 | Yes |
| | WM | PPV | 40 | 0.790 | 0.816 | -0.011 | 0.942 | No |
| | WM | DSC_num | 40 | 0.503 | 0.593 | -0.112 | <0.001 | Yes |
| | WM | Sen_num | 40 | 0.353 | 0.568 | -0.198 | <0.001 | Yes |
| | WM | PPV_num | 40 | 0.844 | 0.792 | 0.046 | 0.010 | Yes |
| | BG | DSC | 40 | 0.277 | 0.333 | -0.064 | <0.001 | Yes |
| | BG | Sen | 40 | 0.166 | 0.213 | -0.067 | <0.001 | Yes |
| | BG | PPV | 40 | 0.811 | 0.819 | -0.008 | 0.681 | No |
| | BG | DSC_num | 40 | 0.507 | 0.602 | -0.042 | 0.043 | Yes |
| | BG | Sen_num | 40 | 0.364 | 0.497 | -0.091 | <0.001 | Yes |
| | BG | PPV_num | 40 | 0.879 | 0.810 | 0.069 | 0.001 | Yes |